\documentclass[journal,twoside,web]{IEEEtran}
\usepackage{generic}
\usepackage{cite}
\usepackage{amsmath,amssymb,amsfonts}
\usepackage{algorithmic}
\usepackage{graphicx}
\usepackage{algorithm,algorithmic}
\usepackage{hyperref}
\hypersetup{hidelinks=true}
\usepackage{textcomp}
\usepackage{subcaption}
\usepackage{makecell}
\usepackage{ragged2e}
\usepackage{tabularx}
\usepackage{array}
\let\labelindent\relax
\usepackage{enumitem}  

\usepackage[T1]{fontenc}
\usepackage[utf8]{inputenc}
\usepackage{lmodern}

\newcommand{\etal}[0]{\textit{et al.}}

\bstctlcite{IEEEexample:BSTcontrol}
\begin{document}

\title{Assessing Physical Frailty and Fall-Risk Indicators with Social Robots: An in situ Evaluation with Older Adults}

\author{Aniol Civit$^{1}$, Antonio Andriella$^{1}$, Alba Martínez$^{2}$, Joan Ars$^{2,3,4}$
, Aida Ribera$^{2}$, Cristian Barrué$^{1}$,  Guillem Alenyà$^{1}$
    \thanks{This work has been submitted to the IEEE for possible publication. Copyright may be transferred without notice, after which this version may no longer be accessible. This work was supported by the project FRAILWATCH 23S06141-001 funded by the Barcelona Council and La Caixa Foundation; the project ROBOCAT SDC006/25/000016 funded by the Generalitat de Catalunya, NextGenerationEU. A. Civit has been supported by AGAUR-FI ajuts (2023 FI-3 00065) Joan Oró of the Generalitat of Catalonia and the European Social Plus Fund.}
    \thanks{$^{1}$Institut de Robòtica i Informàtica Industrial, CSIC-UPC, Llorens i Artigas 4-6, 08028 Barcelona, Spain;
        {\tt\small \{acivit, aandriella, cbarrue, galenya\}@iri.upc.edu}}%
    \thanks{$^{2}$Parc Sanitari Pere Virgili, Area of Intermediate Care, Barcelona, Spain;
        {\tt\small \{jars
        , ariberas\}@perevirgili.cat}}%
    \thanks{$^{3}$RE-FiT Barcelona Research Group, Vall d’Hebron Institute of Research and Parc Sanitari Pere Virgili, Barcelona, Spain}{}
    \thanks{$^{4}$Aging Research Center, Department of Neurobiology, Care Sciences and Society (NVS), Karolinska Institutet and Stockholm University, Widerströmska Huset, Tomtebodavägen 18 A, 171 65, Solna, Stockholm, Sweden}{}
}

\maketitle
\begin{abstract}
Frailty assessments are crucial to evaluate the risk of adverse events and the health and social care needs of older adults, yet their administration remains resource-intensive and typically relies on coarse clinical outcomes, such as task completion times, which may overlook biomechanical indicators of functional decline. To address this, we present a robotic framework that guides older adults through standardised frailty and fall-risk tests while capturing clinical scores and additional frailty-related metrics, offering a deeper insight into a user's condition. 
The system uses a Behaviour Tree architecture that coordinates perception, decision-making, interaction, and measurement modules. Using vision-based skeleton tracking, the robot evaluates established clinical tests, including the Short Physical Performance Battery (SPPB) and the Timed Up and Go (TUG). The framework was co-designed with healthcare professionals and evaluated in situ during six months in a rehabilitation centre's research lab with N=81 older adults. Robot-derived measurements were compared against therapist assessments and clinical reference instruments, including a gait analysis walkway and an inertial measurement unit (IMU). Results showed excellent agreement for most test completion times and gait-related parameters ($ICC > 0.9$). And, substantial agreement for the overall SPPB score comparing the robot and the therapist ($k = 0.67$) and moderate agreement comparing the robot and the IMU ($k=0.55$).
The findings highlight that social robots can provide reliable and objective frailty assessments in healthcare settings while enabling the collection of relevant mobility indicators beyond conventional outcomes. 

\end{abstract}


\section{Introduction}
\label{sec:introduction}

    \begin{figure}[t]
        \centering
        \includegraphics[width=0.95\linewidth]{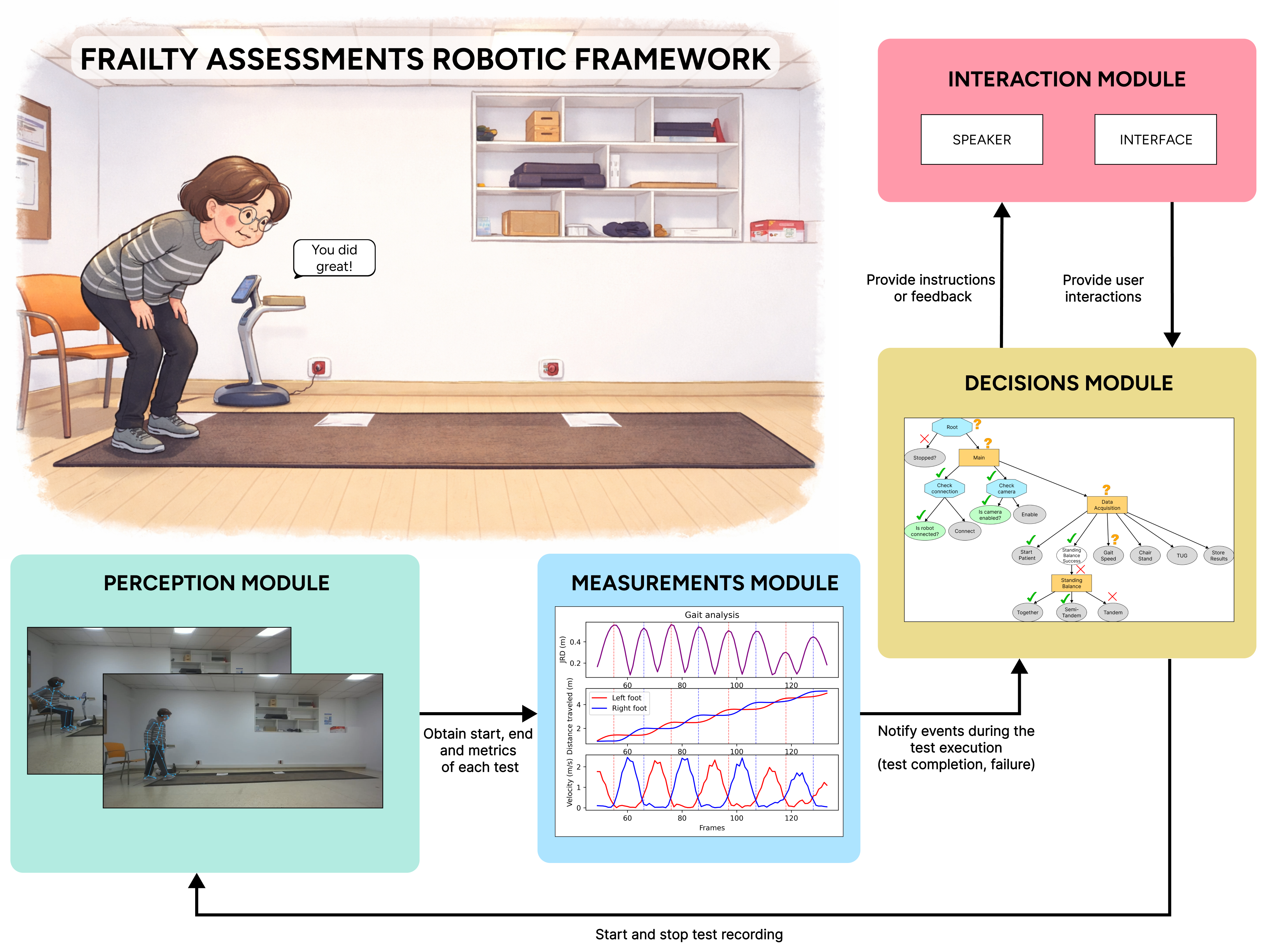}
        \caption{Robotic framework for frailty assessment. The Perception module captures the images from the camera, which are processed by the Measurements module to determine if a test has finished and how the patient performed. That information is used by the Decisions module, which selects which tests to perform at each time. Finally, the decisions trigger the Interaction module, and the robot provides explanations or gives feedback to the patient.}
        \label{fig:framework}
    \end{figure}

    \IEEEPARstart{T}{he} global age distribution is rapidly shifting towards a higher proportion of older adults~\cite{Lam_PDR25}. This socio-demographic transition has started to raise significant concerns, particularly within the healthcare sector~\cite{rudnicka_world_2020}, as older adults often require more specialised care and are more vulnerable to health-related risk factors.
    Current literature indicates that the most reliable predictor of well-being, tolerance to treatment and survival in older adults is frailty status rather than chronological age~\cite{Hamaker_TO12}. Frailty is a geriatric syndrome characterised by a decline in physiological reserve and the organism’s adaptive capacity, resulting in increased vulnerability to stressors and a heightened risk of disability, disease, and mortality~\cite{Fried_TJG01}. 
    
    Assessment models such as the Comprehensive Geriatric Assessment (CGA)~\cite{Lee_JKMS20} and the Integrated Care For Older People Approach (ICOPE)~\cite{Noauthor_book17} offer structured procedures that use validated questionnaires, performance tests, and standardised scales to evaluate multiple dimensions of an older person’s health, including frailty. Both ICOPE and CGA include the assessment of the physical component of frailty, which is achieved by performing the Short Physical Performance Battery (SPPB)~\cite{Guralnik_JG94}. Additionally, the CGA includes the evaluation of basic functional mobility and domains associated with fall-risk through the Timed Up and Go (TUG) test~\cite{Podsiadlo_JAGS91}. These tests consist of simple functional tasks in which the therapist records the completion time to derive a score reflecting the individual’s frailty or risk of falling. However, guiding patients through these assessments requires a fair amount of time from already overburdened healthcare staff~\cite{Bruyere_ACER17}. Moreover, conventional assessments typically rely solely on time-based measurements, while additional objective and frailty-related data could be captured using complementary instruments~\cite{Martinez_JB11, Schwenk_GERONTOLOGY13}. For instance, measuring gait parameters, postural stability during balance tests, or kinematic patterns could provide valuable insights~\cite{Ruiz_SENSORS21}. Previous research in this area has primarily utilised external sensing technologies, such as Inertial Measurement Units (IMUs) or cameras. However, obtaining these additional frailty-related indicators often requires dedicated sensing technologies, specialised equipment, or direct supervision from healthcare professionals, which limits their routine use in clinical practice. 

    To address these challenges, assessment technologies should not only measure clinically relevant metrics accurately but also support autonomous and standardised test administration~\cite{Lachaux_itsg24}. Socially assistive robots offer a promising approach because they can guide individuals through assessment protocols, provide instructions and feedback, and simultaneously collect objective movement data through integrated sensing systems. Recent studies have begun exploring the integration of robots into frailty and fall-risk assessments~\cite{Olde_HT19, Civit_hri24}, as well as the extraction of frailty-related information~\cite{Calabrese_IJSR25}.
    However, these approaches typically lack a comprehensive framework that enables robots to autonomously administer full frailty-risk assessments for older adults or limit their assessment to only time and do not evaluate it with the real end-users.

    We build on our prior work~\cite{Civit_hri24, Civit_roman24}, in which we introduced challenges for robots to assess frailty and risk of falling in older adults and defined methods to measure the performance. In this work, we present a robotic framework to guide and assess such tests autonomously. We evaluate the framework in situ in a healthcare facility over six months with $N=81$ older adults. During the experiments, the robot autonomously guided participants through the tests, under professional supervision to ensure safety, while the measurements obtained by the robot were compared against those recorded by an occupational therapist and reference instruments, namely a walkway mat and an IMU. The results demonstrate that the robot can measure test performance similarly to both the therapist and the reference instruments, highlighting its potential as an effective and objective instrument for assessing frailty and related metrics, while also reducing the workload of healthcare professionals.

    In summary, this work aims to make the following contributions:
    
    \begin{itemize}
        \item We present a robotic framework that administers validated frailty and fall-risk assessments while extracting additional frailty-related biomechanical indicators using vision-based skeleton tracking;
        \item We validate the system in a real healthcare environment through a six-month deployment involving 81 older adults, comparing robot-derived measurements against therapist assessments and clinical reference instruments.
        \item We provide evidence that robotic assessments can mostly achieve substantial agreement with conventional clinical measurements while providing objective biomechanical indicators beyond standard assessment scores.
    \end{itemize}
    

\section{Related Work}

    In this section, we discuss the studies in which frailty assessments are performed. In particular, those performed by using external sensors, such as IMUs, cameras, pressure sensors, grip strength, and others (see Sec.~\ref{sec:related_work_sensors_frailty_assessment}), and those performed by robots (see Sec.~\ref{sec:related_work_robots_frailty}).

    \subsection{Sensor-based frailty assessments}\label{sec:related_work_sensors_frailty_assessment}
    
        A growing body of research has explored the use of sensing technologies to support objective frailty assessment and the extraction of frailty-related indicators beyond conventional clinical outcomes. These approaches typically rely on wearable or environmental sensors.

        Wearable sensing approaches, particularly those based on IMUs, have been widely adopted to assess gait, balance, and lower-limb function. Previous studies have shown that IMU-derived metrics, including gait speed, stride characteristics, velocity profiles, and sit-to-stand dynamics, can discriminate between frailty levels and identify early functional decline~\cite{Martinez_JB11, Millor_JNR13, Martinez_JNR15, Jung_JBHI21, Minici_JBHI22, Eskandari_JBHI22}.
        Other sensing modalities, such as plantar pressure sensors, have also been used to derive gait-related parameters associated with frailty~\cite{Ando_GERIATRICS24}.

        Vision-based systems have emerged as a non-intrusive alternative for measuring frailty-related metrics. Using RGB, RGB-D, or multi-camera configurations, previous studies have extracted gait parameters, postural stability measures, and mobility indicators during standard assessments such as the SPPB and TUG tests~\cite{Gabel_embs12, Yang_SENSORS14, Geerse_PO15, Ginaria_ip16, Gu_wibsn18, Duncan_TBE23}. These approaches demonstrate the potential of markerless motion analysis for objective assessment while avoiding the need for wearable devices.

        \textbf{Research Gaps}: Even though the aforementioned sensors have proven effective, they present some drawbacks for autonomous frailty assessment. IMUs are intrusive and must be correctly positioned on the body; pressure sensors are costly and require controlled environments; and depth-based cameras are limited in range, often requiring a set of multiple cameras. The deep learning models often do not accommodate heterogeneous gait patterns typical of older adults. To our knowledge, no study has autonomously performed frailty assessments independently of participants' physical complexity and without intrusive tools.

    \subsection{Robotic frailty assessments}\label{sec:related_work_robots_frailty}

        Socially assistive robots have increasingly been explored as tools to support the assessment and monitoring of older adults. Their ability to provide embodied guidance, deliver instructions, and interact directly with users makes them a promising platform for administering clinical assessments while simultaneously collecting objective performance measures.

        Previous studies have investigated the use of robots to guide frailty-related evaluations~\etal\cite{Olde_HT19, Palestra_ARR19}. However, they either limited the assessment to a single test~\cite{Olde_HT19} or did not fully disclose the methodology for extracting the additional assessed frailty-related metrics~\cite{Palestra_ARR19}.
        In our previous works, we tried to address such limitations. In~\cite{Civit_hri24}, we conceptualised a framework that enables social robots to administer frailty assessments in healthcare settings. We subsequently validated its measurement capabilities in a controlled study with 22 older adults, comparing robot-derived measures against clinician assessments and an OptiTrack system~\cite{Civit_roman24}. However, the evaluation was limited by the small sample size, controlled experimental conditions, external supervision of the interaction, and the use of a motion-capture system rather than clinical reference instruments.
        Among the most recent developments, Calabrese~\etal\cite{Calabrese_IJSR25} demonstrated that a socially assistive robot can autonomously administer the TUG test while achieving strong agreement with therapist-derived completion times. Lachaux~\etal\cite{Lachaux_roman25} proposed a robotic platform for administering several physical-function assessments, including chair-stand, walking, and grip-strength tests, under healthcare-professional supervision. However, neither study demonstrated autonomous administration of a complete frailty-assessment protocol validated against clinical reference instruments with a statistically relevant sample.

        \textbf{Research Gaps}: Despite the demonstrated potential of robots in frailty assessment, several key gaps remain. Firstly, no study has shown the full implementation of the autonomous robotic behaviour to guide older adults through the frailty and risk of falling assessments. Secondly, there is a lack of validated evidence demonstrating strong agreement between a robot's measurements and those obtained by therapists and/or established reference measurement tools in frailty and functional mobility. This work addresses these limitations by presenting a robotic framework capable of autonomously conducting frailty and fall-risk assessments with older adults, while reliably measuring test performance and some additional frailty-related metrics. 
        
\section{User-Centred Design of Robotic Framework}
\label{sec:user_centered_design_of_robotic_framework}
    
    \subsection{Identify Clinical Needs}
    \label{sec:clinical_needs}  

    The development of the proposed robotic framework followed a user-centred design process carried out in collaboration with healthcare professionals from the Intermediate Care Hospital Parc Sanitari Pere Virgili. Rather than defining the system from a technological perspective, the objective was to understand current clinical practice, identify the main limitations of existing frailty assessment procedures, and derive the functional requirements that the robotic system should satisfy.

    \begin{figure}[t]
            \centering
            \includegraphics[width=1\linewidth]{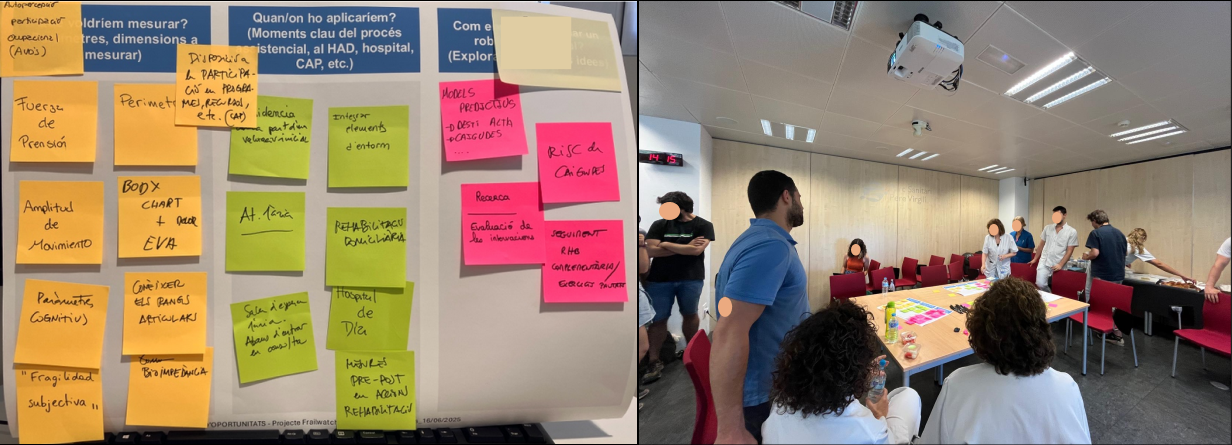}
            \caption{On the left, the answers of a group in the workshop session. On the right, all groups are discussing their answers.}
            \label{fig:co-design}
    \end{figure}

    To this end, we conducted a multidisciplinary co-design workshop involving geriatricians, rehabilitation physicians, physiotherapists, occupational therapists, nurses, and nursing assistants who routinely assess older adults. Participants discussed how frailty assessments are currently performed, the difficulties they encounter in daily practice, and the potential role of a social robot within the clinical workflow (see Fig.~\ref{fig:co-design}).
    
    The discussions revealed three recurrent challenges. First, administering standardised assessments such as the SPPB and the TUG requires dedicated time from healthcare professionals, limiting the frequency with which these assessments can be performed in routine clinical practice. Second, clinical evaluations primarily rely on coarse outcome measures, such as completion times or ordinal scores, while potentially informative biomechanical indicators of mobility and balance are rarely collected because they require additional sensing technologies and increase the complexity of the assessment. Finally, healthcare professionals emphasised the importance of maintaining standardised test administration while reducing their workload, particularly for repetitive screening procedures.
    
    Participants envisioned a system capable of autonomously guiding older adults through standardised assessment protocols, collecting objective measurements of physical performance, and allowing healthcare professionals to focus on interpreting the results rather than administering the tests themselves. Although other assessment domains, including cognition, pain, and activities of daily living, were identified as future opportunities for robotic support, participants consistently prioritised the physical component of frailty as the most suitable starting point due to its standardised procedures and routine use in clinical practice.
    
    These findings established the design requirements for the proposed framework: (i) autonomous administration of standardised physical frailty assessments; (ii) reliable and objective measurement of clinically relevant outcomes; (iii) extraction of additional mobility-related indicators without requiring wearable or dedicated sensing equipment; and (iv) an interaction strategy that is simple, safe, and acceptable for older adults in healthcare environments.

    \subsection{Platform Selection and Initial Robotic Prototype}
    \label{sec:platform_selection_and_initial_robotic_prototype}

    The clinical requirements identified during the workshop guided the development of the first robotic prototype. We focused on implementing the two most widely adopted clinical tests routinely used in geriatric practice: the Short Physical Performance Battery (SPPB) and the Timed Up and Go (TUG).

    To deliver these assessments autonomously, the robotic platform had to satisfy several functional requirements. It needed to provide multimodal instructions through speech and visual cues, detect participant movements without requiring wearable devices, monitor task execution in real time, and ensure patient safety throughout the complete assessment protocol. In addition, the platform had to acquire objective movement measurements beyond the conventional clinical scores, enabling the extraction of additional frailty-related indicators.
    
    Based on these requirements, we selected the Temi\footnote{\url{https://www.robotemi.com/product/temi/}} social robot as the interaction platform. Temi integrates a touchscreen, speakers, microphones, and an autonomous mobile base, making it well-suited for guiding participants through standardised assessment procedures. Because onboard computational resources were insufficient for real-time vision-based skeleton tracking, a stereoscopic ZED2i camera connected to an external GPU-equipped computer was integrated into the system to provide full-body motion capture throughout the assessments.
    
    This first prototype established the core capabilities required for autonomous frailty assessment. However, before deployment in a clinical environment, the interaction flow, assessment protocol, and measurement procedures were iteratively refined together with healthcare professionals to ensure their suitability for routine clinical practice.

    \subsection{Iterative Co-design and Prototype Refinement}
    \label{sec:interative_co_design_and_prototype}

    Following the implementation of the initial prototype, the system underwent several cycles of iterative refinement in collaboration with a multidisciplinary team comprising a physiotherapist, an occupational therapist, and a neuropsychologist. Through regular co-design sessions, healthcare professionals evaluated the interaction flow, assessed the suitability of the clinical protocol, and identified situations that could compromise either the validity of the assessments or the safety of the participants.
    
    These sessions highlighted several important design considerations. First, participants may misunderstand standardised test instructions, particularly when performing the assessments autonomously. Consequently, the robot was designed to provide multimodal guidance by combining verbal explanations with visual demonstrations of each task. Second, healthcare professionals emphasised that assessment scoring should not be influenced by excessive encouragement or corrective feedback during test execution. Therefore, the robot limits its interaction while participants perform the tests and provides feedback only after each assessment has been completed. Third, several safety-related situations were identified, including incorrect foot placement during the Standing Balance test, incomplete chair stand movements, and the need for participants with limited balance to perform the assessments with appropriate supervision. These observations guided both the interaction design and the measurement algorithms implemented in the final framework. In particular, given the safety concerns during the assessments, the healthcare professionals agreed that the robot should be used only with professional supervision until further testing confirms its safety across the target patient population.
    
    Before the main study, the refined prototype was evaluated in a pilot deployment involving four older adults in the physiotherapy laboratory, where the final evaluation would take place. The pilot focused on assessing the usability of the interaction, the suitability of the experimental setup, and the reliability of the vision-based measurements under realistic conditions.
    
    The pilot resulted in three final modifications to the system. First, the stereoscopic camera was repositioned perpendicular to the walking direction to improve skeleton tracking accuracy during the Chair Stand and TUG assessments. Second, the Standing Balance instructions were revised to explicitly encourage participants to look forward instead of toward the robot's display while maintaining the balance position. Third, verbal feedback during the Chair Stand repetitions was removed to avoid influencing participants' performance. Finally, to ensure safety and address any issues due to potential miscommunication between the participants and the robot, we decided to have an occupational therapist supervise the whole evaluation.
    
    The resulting prototype was considered sufficiently robust for long-term deployment in the clinical study and constitutes the robotic framework evaluated in the remainder of this paper.

\section{Robotic Framework for Autonomous Frailty Assessments}

    In this section, we present the complete design of the robotic framework used to autonomously conduct frailty and fall-risk assessments. The framework integrates all refinements derived from the co-design process and pilot iterations described in Section~\ref{sec:user_centered_design_of_robotic_framework}.

    The architecture of the system is illustrated in Fig.~\ref{fig:framework}. The framework is composed of four main modules. First, the Perception Module processes the sensory input and estimates the participant’s skeletal pose using a vision-based tracking system (see Sec.~\ref{sec:perception-module}). Second, the Decision-Making Module coordinates the assessment procedure via a Behaviour Tree (BT)~\cite{Iovino_RAS22}, which determines the robot's actions based on the current assessment state and the participant’s performance (see Sec.~\ref{sec:behaviour_tree}). Third, the Interaction Module communicates instructions and feedback to the participant through a graphical interface and verbal prompts (see Sec.~\ref{sec:interaction_interface}). Finally, the Measurement Module processes the recorded motion data to compute the test completion times and extract frailty-related metrics (see Sec.~\ref{sec:measurement_algorithms_and_metrics}).

    \subsection{Perception Module}\label{sec:perception-module}
    The perception module is responsible for capturing and processing visual information to estimate the participant’s body pose during the frailty assessments in real-time. 

    The system relies on a stereoscopic depth camera (ZED2i) to observe the participant during the assessments. The camera was selected for its ability to reliably estimate 3D human skeletons at distances up to 10 meters, which is required to capture the participant during walking-based tests such as the Gait Speed and TUG assessments. The camera is positioned at a distance that keeps the participant within the field of view throughout the entire walking trajectory (see Fig.~\ref{fig:lab_distribution}). This configuration enables the system to monitor both static tests, such as the Standing Balance and Chair Stand assessments, and dynamic walking tasks.
    
    The camera is directly connected to an external computer, while communication between the robot and the computer is established through a WebSocket interface. This architecture enables the robot to trigger perception processes and retrieve motion data during the execution of the BT controlling the assessment.
    
    Human pose estimation is performed using the skeleton tracking module provided by the ZED SDK. The system estimates the 3D positions of 38 body keypoints representing the main human joints, including the head, torso, pelvis, hips, knees, ankles, and feet. These keypoints form a skeletal representation of the participant that allows the framework to monitor body posture and detect relevant movements during the assessments.
    
    During the execution of each test, the detected 3D joint coordinates are continuously streamed and stored in a temporal buffer. These data are used to detect movement events and compute biomechanical metrics in later stages of the framework. To reduce noise in the estimated skeleton, the joint coordinates are filtered before further processing. The resulting skeletal representation provides the spatial information required to monitor the participant’s posture and movement throughout the frailty assessment.

    \subsection{Decision-making Module}\label{sec:behaviour_tree}

\begin{figure}[t]
            \centering
            \includegraphics[width=0.9\linewidth]{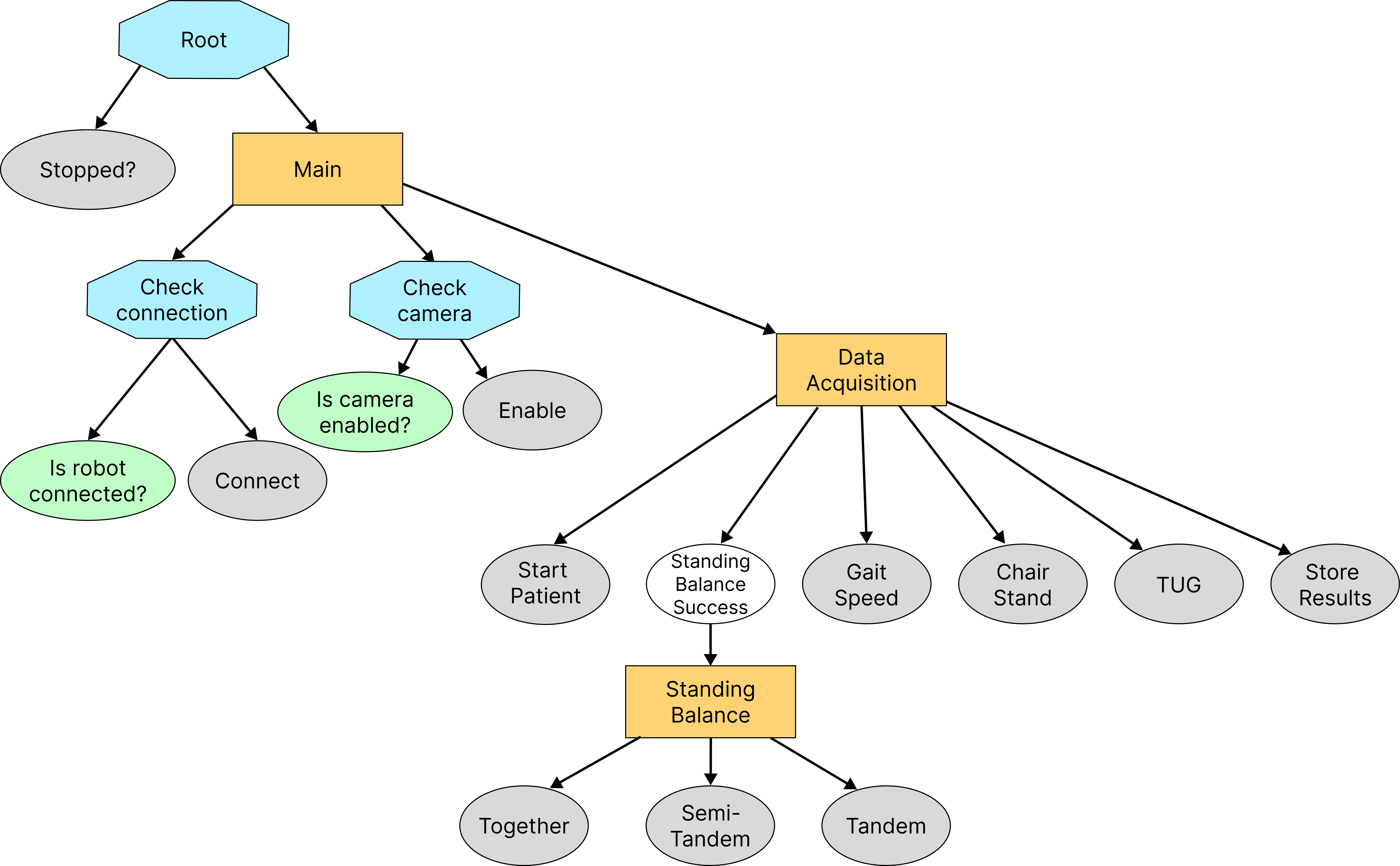}
            \caption{Behaviour Tree (BT) for the Frailty Assessment, the blue octagons are selector nodes, the orange rectangles are sequential nodes, the grey circles are action nodes, the green circles are condition nodes, and the white circle is a decorator of type ``Failure is Success'', called ``Standing Balance Success'', to guarantee that when the participant fails in one of the balance tests, the rest of the balance tests are skipped and the robot moves to the Gait Speed test. Each test is a robot action of the BT.}
            \label{fig:behaviour_tree}
        \end{figure}

        The decision-making module coordinates the execution of the frailty assessment and determines the actions performed by the robot throughout the interaction with the participant. The control architecture is implemented using a BT, a hierarchical decision-making structure widely used in robotics due to its modularity, robustness, and ease of extension~\cite{Gugliermo_RAS24, Merlo_TRO25}. 
        The BT orchestrates the complete assessment procedure by sequencing the different tests and coordinating the interaction with the participant. In addition, it integrates information provided by the perception module in order to detect relevant events during the execution of the tests. The structure of the BT used in this work is illustrated in Fig.~\ref{fig:behaviour_tree}.

        At the beginning of the session, the BT executes a set of system checks to ensure that all required hardware components are available. In particular, the system verifies the connection between the robot and the external computer responsible for skeleton tracking, as well as the availability of the stereoscopic camera used for motion capture. If these conditions are satisfied, the robot introduces itself to the participant and explains the purpose of the assessment.

        After the introduction, the BT guides the participant through the complete physical frailty evaluation procedure, which includes the Standing Balance, Gait Speed, Chair Stand, and Timed Up and Go tests.

        Each physical test is implemented as an action node within the BT and follows a similar execution structure. The robot first explains the procedure with verbal and visual instructions to the participant. Once the participant confirms readiness, the perception module begins monitoring the participant's movements while detecting relevant events, such as the start and completion of movements required for each test. After detecting such conditions, the corresponding action node returns success, allowing the Behaviour Tree to transition to the next test.
        
        The BT determines when a test has been completed or when a failure condition occurs. For instance, during the Standing Balance assessment, if an imbalance is detected in the Together or Semi-Tandem positions, the test is terminated according to the SPPB protocol, and the BT transitions to the next assessment (Gait Speed). This behaviour is implemented using a decorator node that converts the failure of a balance position into a successful completion of the balance test, ensuring that the remaining balance positions are skipped.
        
        Once a test is finished, the robot provides feedback to the participant, and the BT transitions to the next assessment node. After all tests have been performed, the BT terminates the evaluation process and stores the collected data for subsequent analysis (extracting times and additional metrics).

    \subsection{Interaction Module}\label{sec:interaction_interface}

        Interaction between the participant and the robot was primarily conducted through the robot’s touchscreen interface, which served as the main communication channel. Speech recognition was not implemented, as automatic speech recognition systems often exhibit reduced accuracy when used by older adults due to age-related vocal changes~\cite{Werner_hfesam19}. Consequently, participants interacted with the robot exclusively through touch-based inputs.

        At the beginning of the interaction, the robot’s screen displays a simple animated face together with two buttons: Start, which initiates the assessment sequence, and Language, which allows participants to select the interaction language. The available languages were \textit{Catalan} and \textit{Spanish}. 

        During the physical tests, the robot provides instructions through a combination of visual and verbal cues. For the Standing Balance test, the robot displays an image illustrating the correct foot position while providing a verbal explanation of how to perform it. After the explanation, a countdown is initiated when the measurement starts. For the Gait Speed, Chair Stand, and TUG tests, the robot presents a video demonstration together with a verbal explanation of the procedure. Beneath the video, two buttons are available: ``Start'', to begin the test, and ``Repeat'', to replay the instructions. In the Standing Balance test, the ``Repeat'' button is intentionally omitted to minimise the risk of falls, as participants typically position their feet during the explanation and take some time to remain stable in those positions, and reaching toward the robot's screen in that stance could compromise their stability. Upon completion of all assessments, the robot announces the end of the experimental session.

        All robot voice dialogues are pre-recorded Text-To-Speech (TTS) audio files obtained using Microsoft Azure, both in \textit{Catalan} and \textit{Spanish}. The dialogues were revised and modified during the pilot studies to achieve better guidance for the tests. The reasons for using pre-recorded audio files are that: the pre-recorded provided more control over what the robot says, avoiding unsafe or wrong messages, and that most scripts are standardised due to test restrictions. The test explanations are also standardised\footnote{Complete SPPB test: https://geriatrictoolkit.missouri.edu/SPPB-Score-Tool.pdf}.

    \subsection{Measurement Module}\label{sec:measurement_algorithms_and_metrics}

        The measurement module processes the skeletal data provided by the perception module to detect test events and extract performance metrics during the frailty assessments. Using the estimated 3D joint coordinates, the system automatically determines the start and end of each test and computes both completion times and additional biomechanical metrics associated with frailty and mobility performance.

        During the execution of each assessment, the skeletal coordinates are continuously analysed in real time. Specific movement patterns are used to identify relevant events such as standing up, initiating gait, or completing a walking trajectory. These events enable the system to measure the duration of each test (see Sec.~\ref{sec:test_completion}) and derive quantitative indicators of physical performance (see Sec.~\ref{sec:additional_variables_measurements}).

        \subsubsection{Test completion times}\label{sec:test_completion}
        
            The measurement of test completion times follows the approach proposed in our previous work~\cite{Civit_roman24}. In this method, the start and end points of each test are automatically detected according to the following conditions:

            \begin{itemize}
                \item Standing Balance: The test begins immediately after the robot's countdown. The test ends either when the participant maintains the required position for 10 seconds or when a loss of balance is detected. 
                \item Gait Speed: The test begins when the participant presses the ``Start'' button. To obtain steady-state consistent walking gait information, we include an acceleration phase of 1 meter. Time measurement starts once the participant has crossed 1 meter from the initial position, and it ends when the participant reaches the 5-meter mark along the walking trajectory. 
                \item Chair Stand: The test begins when the participant transitions from a seated to a standing posture. Joint angles between the torso, hips, and knees are monitored to detect sit-to-stand and stand-to-sit transitions. The test ends after five complete stand-up repetitions are detected.
                \item TUG: The test starts when the participant first stands up from the chair and ends when the participant returns to the seated position after completing the walking trajectory.  
            \end{itemize}
            
            To detect imbalances during the Standing Balance test, the algorithm calculates the distances between the toes of each foot and the heels, monitoring for significant variations within a sliding time window. For the Chair Stand and TUG tests, knee and hip angles are calculated by tracing vectors between the shoulder, hip, knee, and ankle keypoints. These angles allow the algorithm to determine whether the participant is sitting, standing, or transitioning between the two states.

        \subsubsection{Frailty-related additional metrics}\label{sec:additional_variables_measurements}

            The additional frailty-related metrics build upon and extend the approach proposed in our previous work~\cite{Civit_roman24}. These metrics include those measured by the reference instruments used in the experimentation, namely the IMU and the walkway mat. The metrics are:

            \begin{itemize}
                \item \textbf{Step Length}: Longitudinal distance between the positions of both feet after a step.
                \item \textbf{Stride Length}: The longitudinal distance between two consecutive steps of the same foot.
                \item \textbf{Stride Velocity}: The longitudinal velocity between consecutive steps of the same foot.
                \item \textbf{Chair Stand Ascension Inclination Range}: The angular range through which the participant tilts their torso forward before standing up. This variable helps determine whether the participant relies primarily on body momentum or lower-limb muscle strength to rise. 
                \item \textbf{Chair Stand Ascension Peak Acceleration}: The maximum acceleration of the pelvis, approximating the participant's centre of gravity, along the direction of movement while standing up.
                \item \textbf{Chair Stand Ascension Peak Power}: The maximum power generated while standing up, calculated as $P=F_a \cdot v_a$, where $F_a$ is the force and $v_a$ is the velocity at the same instant.
                \item \textbf{Chair Stand Descension Impulse}: The maximum acceleration recorded during the sitting phase of the test. 
                \item \textbf{Balance Eccentricity}: Quantifies whether the torso displacement occurs predominantly in one direction or is evenly distributed. It is computed as $\mathcal{E} = \sqrt{1-\frac{b^2}{a^2}}$, where a$>$b are the major and minor axes of the fitted ellipse, respectively. The value ranges from 0 to 1: values near 0 indicate circular movement, while values near 1 indicate an elongated elliptical trajectory.
                \item \textbf{Balance Stability}: The area of the ellipse encompassing the trajectory of the upper torso inclination during the Standing Balance test. 
                \item \textbf{Balance Direction}: The predominant angle of upper torso inclination during the Standing Balance test, indicating the main direction of postural sway.
            \end{itemize}

\section{Experimental Design}\label{sec:experimentation}
    This section describes the experimental methodology used to evaluate the proposed robotic framework. The objective of the study is to assess whether the robot can autonomously guide older adults through frailty assessments and obtain measurements comparable to those obtained by a therapist and clinical reference instruments. 
    As already mentioned, an occupational therapist supervised the experiment for safety and to intervene if something unexpected occurred.

    Agreement between quantitative metrics was evaluated using the ICC(2,1)~\cite{Koo_JCM16}, the two-way random model for single measures, which measures reliability based on consistency rather than absolute agreement, assuming the raters are fixed. Categorical scores were compared using weighted Cohen's Kappa coefficient ($\kappa$). According to standard interpretation, ICC values below 0.5 indicate poor reliability, values between 0.5 and 0.75 indicate moderate reliability, values between 0.75 and 0.9 indicate good reliability, and values above 0.9 indicate excellent reliability. Similarly, $\kappa$ values below 0 indicate poor agreement, values between 0 and 0.2 indicate slight agreement, 0.2-0.4 indicate fair agreement, 0.4-0.6 indicate moderate agreement, 0.6-0.8 indicate substantial agreement, and 0.8-1.0 indicate almost perfect agreement.
    
    \subsection{Apparatus}\label{sec:apparatus}

        The experiments were conducted using a Temi robot as discussed in Sec.~\ref{sec:platform_selection_and_initial_robotic_prototype}. Since skeleton tracking requires GPU acceleration, a ZED2i stereoscopic camera was connected to an external computer equipped with an NVIDIA GeForce RTX 4080 GPU. 

        Two reference instruments were used for comparison: the GaitRITE walkway mat~\cite{Bilney_GP03}, which measures gait parameters during walking tasks, and an Xsens IMU~\cite{Roetenberg_XSENS09} placed on the participant’s lower back to capture upper-body motion during the tests.

    \subsection{Hypotheses}\label{sec:hypotheses}

        We evaluated the following hypotheses:
    
        \begin{enumerate}[leftmargin=3em, labelwidth=!, itemindent=0pt, widest=9]
        \item[H1a:] The test completion times and frailty scores measured by the robot show strong agreement with those measured by the therapist.
        \item[H1b:] The test completion times and frailty scores measured by the robot show strong agreement with those measured by the reference instruments.
        \item[H2:] The additional frailty-related metrics extracted by the robot show strong agreement with those obtained using reference instruments.
        \end{enumerate}


    These hypotheses build upon our previous work~\cite{Civit_roman24}, in which we validated the proposed metrics against clinician assessments and used the OptiTrack motion capture system as the ground truth to evaluate both test completion times and frailty-related metrics in a smaller participant cohort. In the present study, we evaluate the robotic system in a real-world clinical setting using a statistically significant sample size and compare its measurements with those obtained by a therapist and by standard reference instruments commonly used in clinical practice.

    Finally, we formulate an additional exploratory hypothesis (H3) regarding the robot's ability to administer the assessments autonomously. Specifically, we evaluate whether therapist intervention is required during test administration and, if so, to what extent. aim to evaluate whether and to what extent therapists need to intervene during the test performance.
        
    \subsection{Inclusion and Exclusion Criteria}\label{sec:inclusion_exclusion}

        Participants were eligible for inclusion if they met the following criteria:  
        (i) age equal to or greater than 65 years;  
        (ii) presence of at least one indicator associated with potential frailty from those described by ICOPE~\cite{Noauthor_book17} such as subjective memory complaints, depression, balance problems, having fallen twice within the previous six months, weight loss, fatigue, strength loss, reduced functional capacity, or severe comorbidity; and (iii) sufficient cognitive and physical capacity to interact with the robot and provide informed consent.
        
        Participants were excluded if they met any of the following conditions: need for hospital admission due to a condition that precludes safe participation or compliance with study procedures, or a life expectancy of less than one year.
        
        Participants retained the right to withdraw from the study at any time after providing consent. In such cases, their participation was terminated immediately, and their data were excluded from the analysis without requiring justification or causing any negative consequences.

    \subsection{Participants}\label{sec:participants}
        

        The sample size was determined by two a priori precision-based calculations for the primary reliability endpoints.

        For continuous metrics (test completion times and frailty-related metrics), agreement was quantified with the Intraclass Correlation Coefficient (ICC). Applying Bonett's formula~\cite{Bonett_SM02} for a 95\% confidence interval of width 0.20 around an expected ICC between 0.75 and 0.8 with 2 raters (pairwise comparisons) yielded a required sample size between 50 and 74 observations. 
        For categorical metrics (individual test scores [0 to 4] and total score [0 to 12]), the agreement was computed with Cohen's Kappa.
        Following Bujang and Baharum's framework~\cite{Bujang_EBPH17}, with a 95\% confidence interval of width 0.20, with a null value of 0.50 (substantial agreement per Landis and Koch~\cite{Landis_B77}), and an expected Kappa between 0.7 and 0.8, the required sample was between 28 and 69 for 5-category scores. The sample size for more categories is larger, but since the SPPB score is computed from the sum of the others, this can be dismissed. 
        
        A total of 81 older adults were recruited for the study, anticipating a 10\% of dropout to reach the minimum sample size required, with 4 additional participants from the pilot study, whose results are not included in the analysis. The mean age of the participants was 75.06 years (SD = 7.21). The sample consisted of 18 male and 63 female participants, meaning it was not evenly distributed by gender.
        Participant recruitment was compromised by the fact that they were older adults. They had to be willing to come to the hospital (which is in a steep part of Barcelona), sometimes only to participate in the study. 
    
        Not all participants were able to complete every physical test due to physical limitations. In total, 76 participants completed the Gait Speed test, 75 completed the Timed Up and Go (TUG) test, and 67 completed the Chair Stand test. According to the Short Physical Performance Battery (SPPB) protocol, when a participant was unable to perform a test, the corresponding clinical score was assigned a value of zero. However, the associated test duration was excluded from time-based analyses because no valid completion time could be recorded. Additionally, one observation from the IMU in the Chair Stand was removed from the study since it provided a test completion time of 3 seconds, which is not realistic, and disagreed by more than 10 seconds with the therapist's measure. 
    
        Participants were recruited from several sources, including hospitalised patients undergoing recovery, day-care centre attendees, nursing home residents, and others. This recruitment strategy was designed to obtain a heterogeneous sample including both frail and non-frail older adults.

    \subsection{Study approval}\label{sec:study_approval}

        This research study has been approved by the Clinical Research Ethics Committee (CEIm), Vall d’Hebron University Hospital, Barcelona, Spain, with ethics board protocol number PR(PV)66/2024 the 13th of September of 2024. 
        
    \subsection{Experimental Setting}\label{sec:experimental_setting}

        The experiments were conducted in the physiotherapy laboratory of a hospital. The laboratory consists of two rooms: a reception room used to welcome participants, explain the study, and obtain informed consent, and a second room where the robotic assessments were performed.

\begin{figure}[t]
            \centering
            \includegraphics[width=0.9\linewidth]{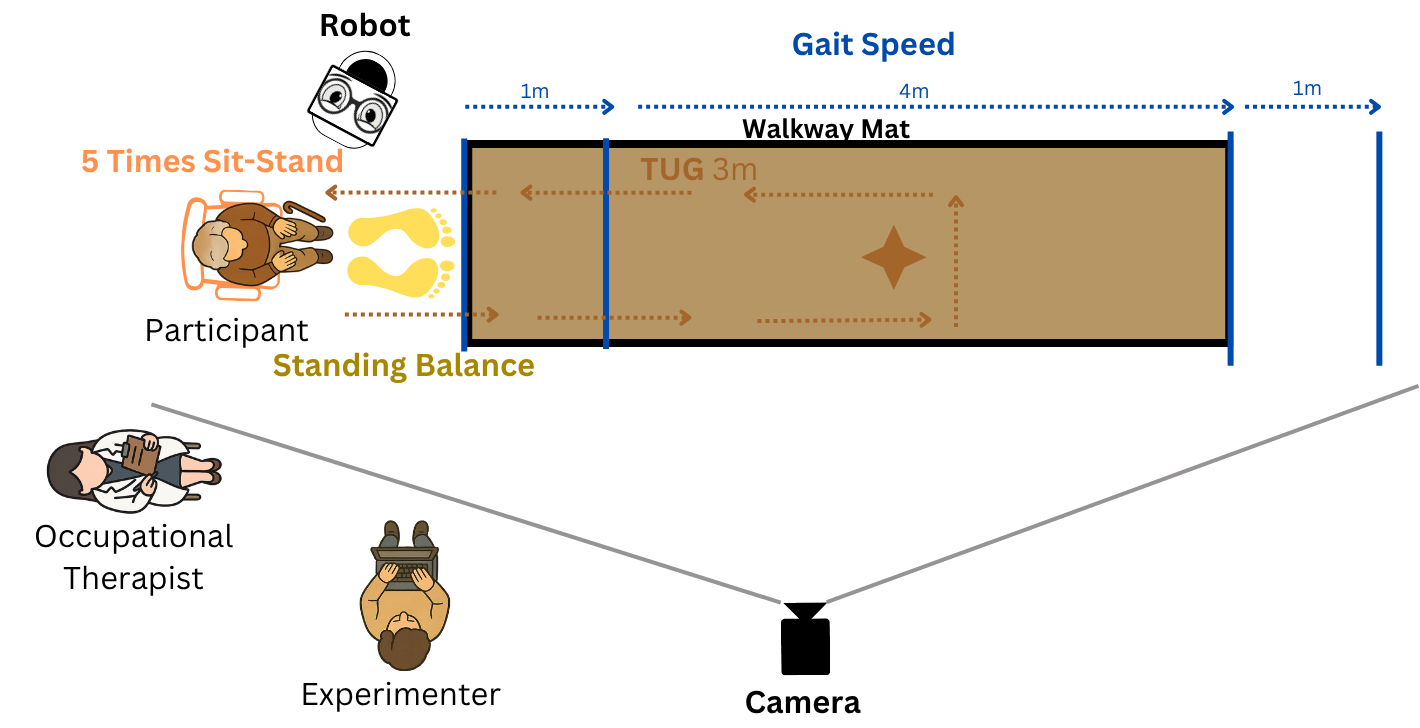}
            \caption{Experimental setting of the frailty assessment experiment.}
            \label{fig:lab_distribution}
        \end{figure}

        The experimental setup is illustrated in Fig.~\ref{fig:lab_distribution}. Participants began each assessment seated on a chair positioned in front of the robot. A 5-meter walkway mat was placed in front of the participant to record gait parameters during the walking tests. All assessments started from the same location, marked on the floor next to the robot.
        
        The Gait Speed test was performed along the walkway mat. An additional meter beyond the mat allowed participants to decelerate naturally after completing the test. For the TUG test, a mark located 3 meters from the chair indicated the turning point of the walking trajectory. The Standing Balance and Chair Stand tests were performed outside the mat area.
        
        The stereoscopic camera used for skeleton tracking was positioned at a distance that ensured full-body visibility during the walking tests while maintaining accurate skeletal tracking throughout the assessment. 
        
        During the assessments, the therapist remained close to the participant to ensure safety. Both the therapist and the experimenter stayed outside the camera's field of view to avoid interfering with the skeleton tracking measurements.

    \subsection{Protocol}\label{sec:protocol} 

        Upon arrival at the laboratory, participants were welcomed by the occupational therapist, who explained the study and obtained written informed consent. Participants then provided basic demographic information. The therapist subsequently attached the IMU to the participant's lower back and activated both reference instruments: the IMU and the walkway mat.

        After the sensors were configured, the robot-guided frailty assessment began. Participants remained seated on the chair positioned in front of the robot and initiated the session by pressing the \textit{Start} button on the robot's interface. The robot then guided the participants through the physical tests.
         
       The therapist manually started and stopped the recordings of the IMU and walkway mat at the beginning and end of each test, and measured the completion times using a chronometer. If a test was performed incorrectly or a technical issue occurred (e.g., sensor disconnection), the therapist requested that the participant repeat the test.
        
        After completing all assessments, the participant was escorted out of the evaluation room.

    \subsection{Tests Evaluation Measures}\label{sec:evaluation_measures}

        To address H1, the completion times measured for each SPPB subtest were converted into clinical scores according to the standard SPPB scoring thresholds.
        
        \textbf{Standing Balance.} Participants were asked to maintain balance for up to 10 seconds in progressively more challenging foot positions: Together, Semi-Tandem, and Tandem. Successfully maintaining the Together and Semi-Tandem positions granted 1 point each and allowed progression to the next position. In the Tandem position, participants received 2 points if the position was maintained for the full 10 seconds, or 1 point if maintained between 3 and 10 seconds. If balance was lost during the Together or Semi-Tandem positions, the test ended, and a score of 0 was assigned for that position.

        \textbf{Gait Speed.} Participants walked a distance of 4 meters. The completion time was converted into a score according to the SPPB thresholds: 4 points for times under 4.82 seconds, 3 points for 4.82–6.21 seconds, 2 points for 6.21–8.7 seconds, 1 point for 8.7–60 seconds, and 0 points if the test was not completed or exceeded 60 seconds.

        \textbf{Chair Stand.} Participants performed five consecutive chair stands. The score was determined according to the completion time: 4 points for times under 11.19 seconds, 3 points for 11.19–13.69 seconds, 2 points for 13.69–16.69 seconds, 1 point for 16.69–60 seconds, and 0 points if the test was not completed or exceeded 60 seconds.

        The three subtest scores were summed to compute the overall SPPB score, which ranges from 0 to 12 points. Scores are used for screening. Scores between 0 and 6 indicate potential frailty, between 7 and 9 indicate pre-frailty, while higher scores indicate normal physical function~\cite{Pritchard_Geriatrics17}.

        Risk of falling was evaluated using the TUG. Participants completing the test in less than 10 seconds were considered to have good mobility, those requiring between 10 and 20 seconds were classified as having regular mobility, and those exceeding 20 seconds were considered to have an increased risk of falling.

        In addition to the primary clinical scores, to provide evidence for H2, several biomechanical metrics were analysed to further characterise participants' physical performance (see Sec.~\ref{sec:additional_variables_measurements} for more detail). The metrics obtained from the IMU included: Chair Stand Ascension Inclination Range, Chair Stand Ascension Peak Acceleration, Chair Stand Ascension Peak Power, Chair Stand Descension Impulse, Balance Eccentricity, Balance Stability, and Balance Direction. The gait-related metrics obtained from the walkway mat included Left and Right Step Length, Stride Length, and Stride Velocity.

        Finally, to investigate our explorative hypothesis H3, the autonomy of the robotic framework was evaluated by annotating the number of interventions required from the therapist during the assessments and the reasons for these interventions.

    \section{Results and Discussion}

        This section presents the results obtained from the in situ evaluation of the proposed robotic framework. The analysis focuses on assessing the agreement between the metrics obtained by the robot and those recorded by the therapist and reference instruments.


        \subsection{H1a: Agreement Between Robot and Therapist Measurements}\label{sec:exp_times_doctor_robot}

\begin{figure*}[t]
    \centering
    \begin{subfigure}{0.19\linewidth}
        \centering
        \includegraphics[width=\linewidth]{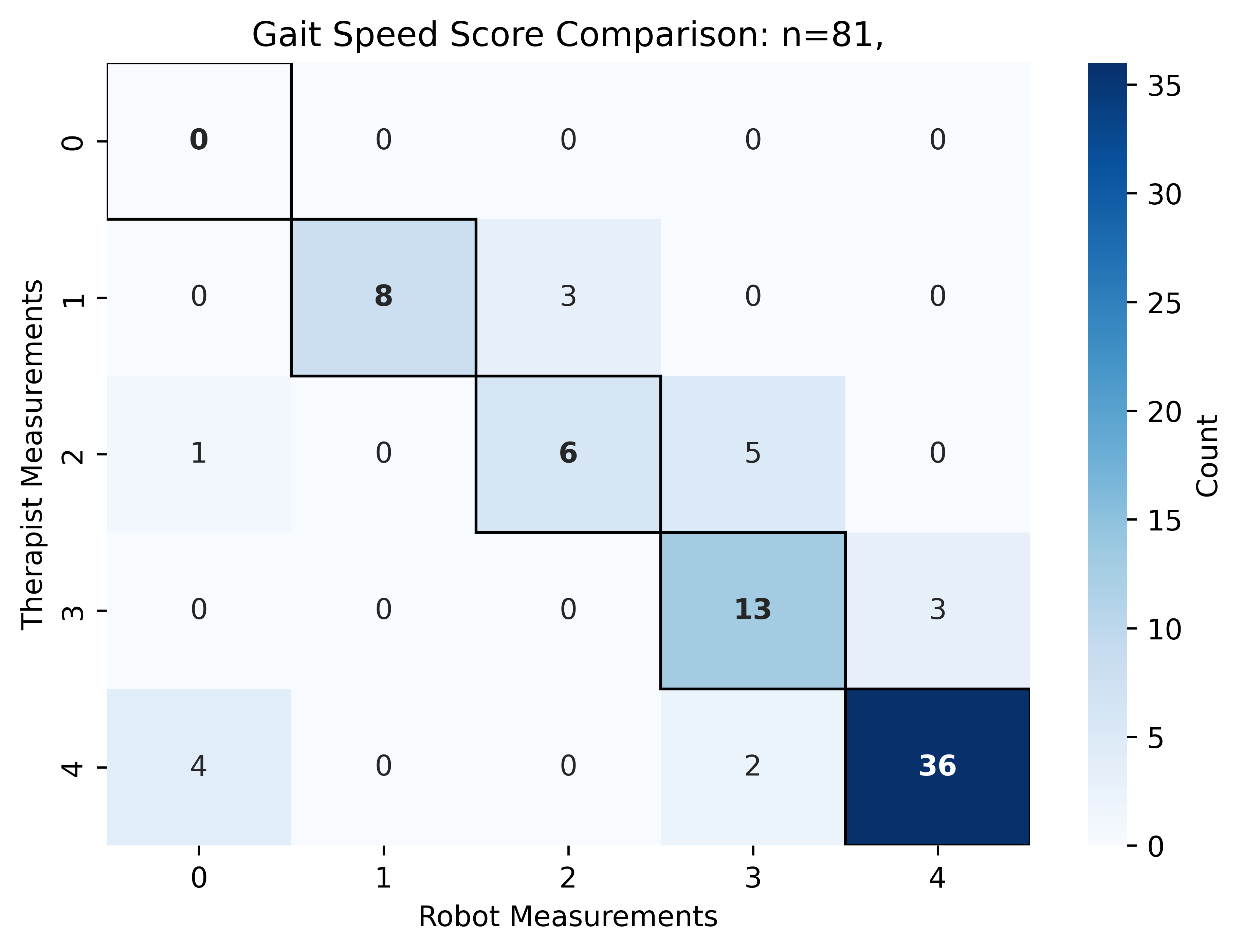}
        \caption{Gait Speed}
        \label{fig:robot_therapist_gait_speed_score}
    \end{subfigure}\hfill
    \begin{subfigure}{0.19\linewidth}
        \centering
        \includegraphics[width=\linewidth]{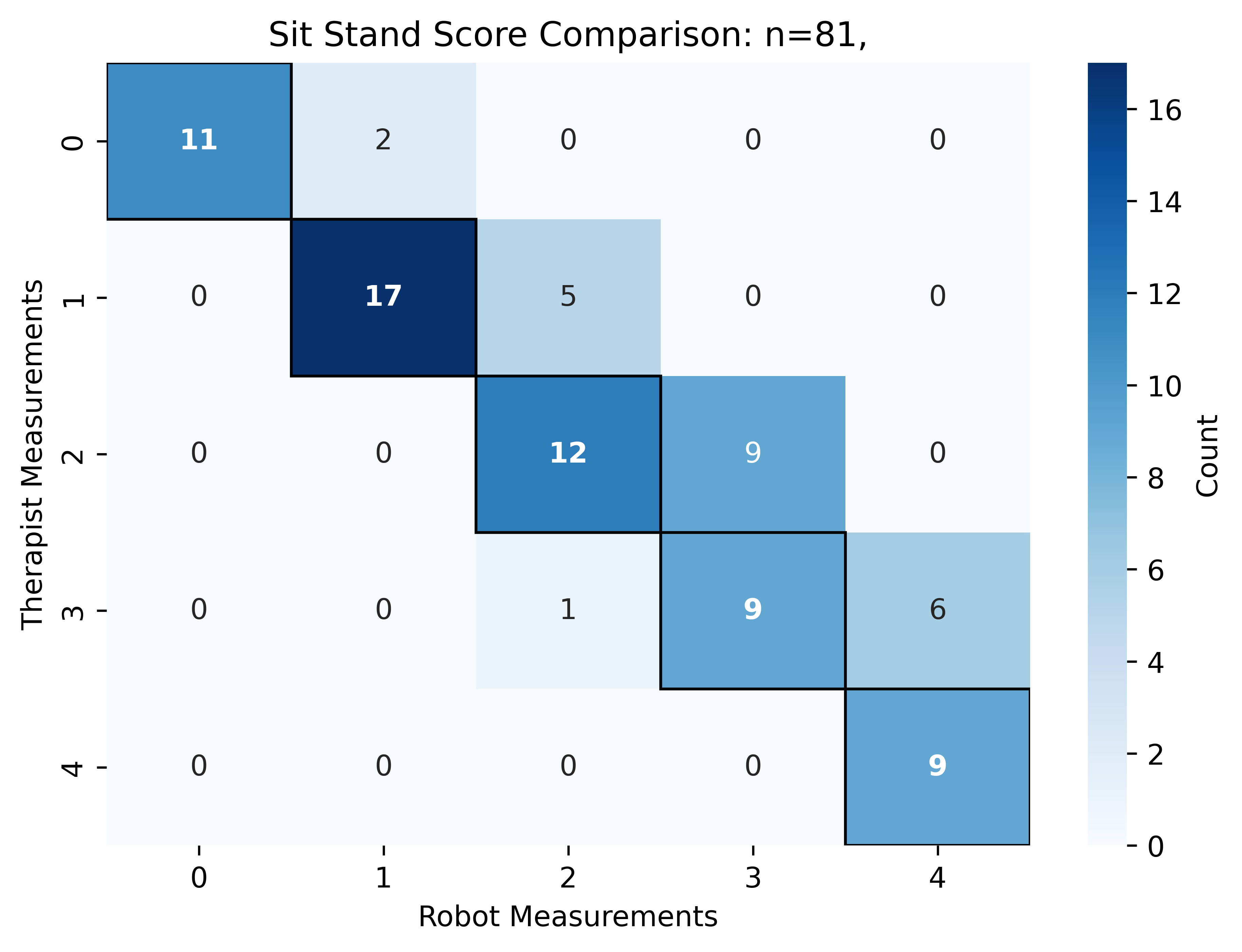}
        \caption{Chair Stand}
        \label{fig:robot_therapist_sit_stand_score}
    \end{subfigure}\hfill
    \begin{subfigure}{0.19\linewidth}
        \centering
        \includegraphics[width=\linewidth]{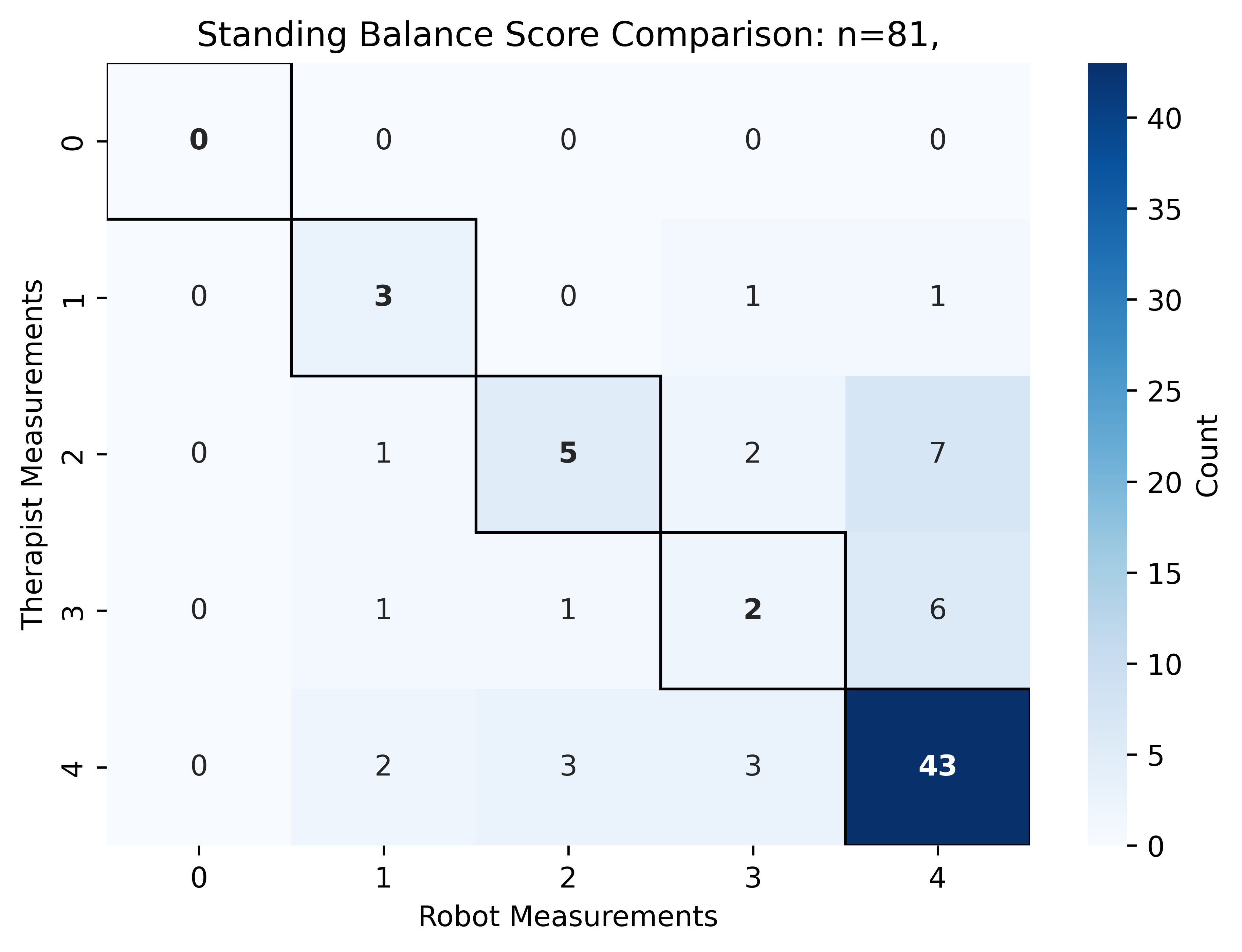}
        \caption{Standing Balance}
        \label{fig:robot_therapist_standing_balance_score}
    \end{subfigure}\hfill
    \begin{subfigure}{0.19\linewidth}
        \centering
        \includegraphics[width=\linewidth]{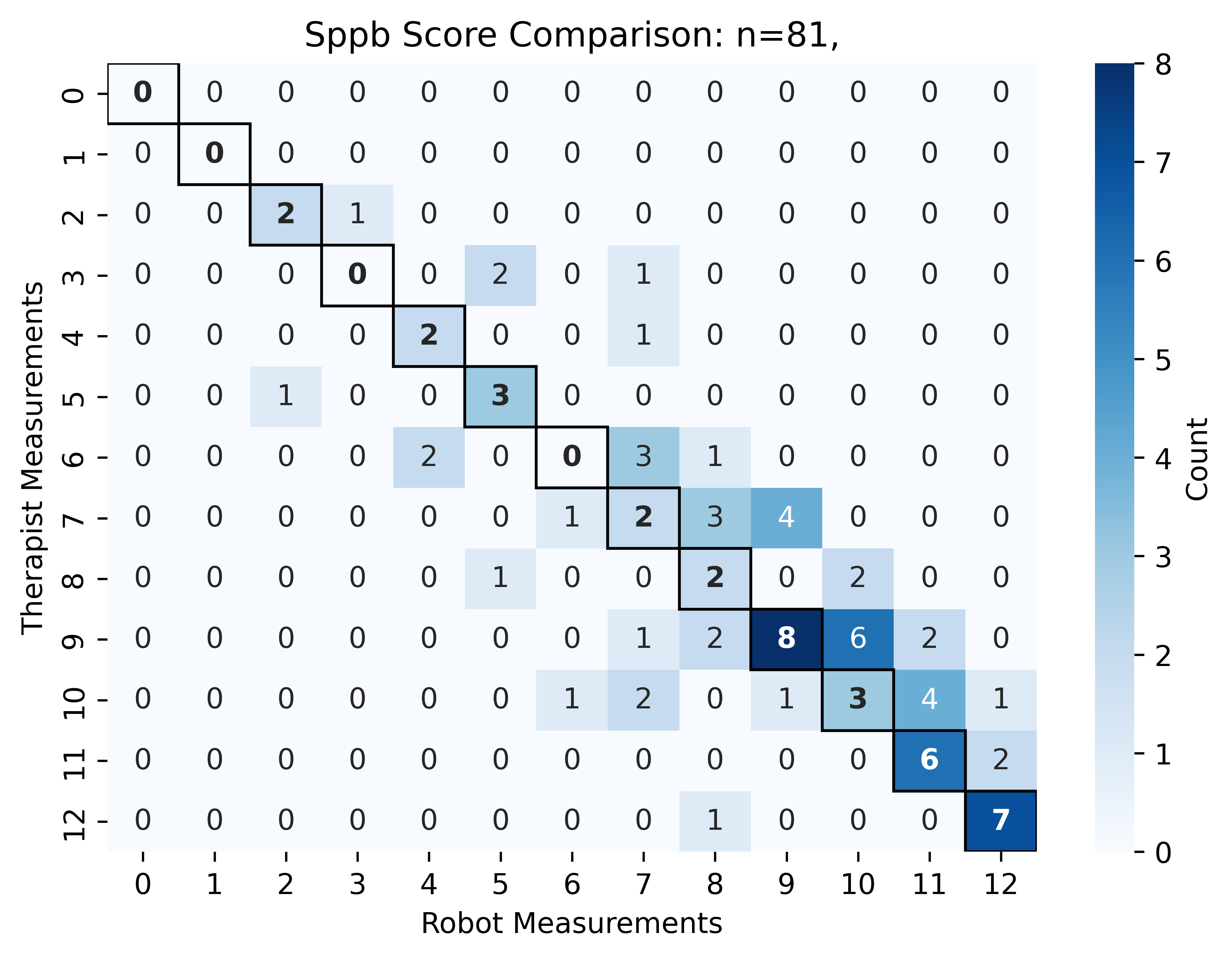}
        \caption{Full SPPB}
        \label{fig:robot_therapist_sppb_score}
    \end{subfigure}\hfill
    \begin{subfigure}{0.19\linewidth}
        \centering
        \includegraphics[width=\linewidth]{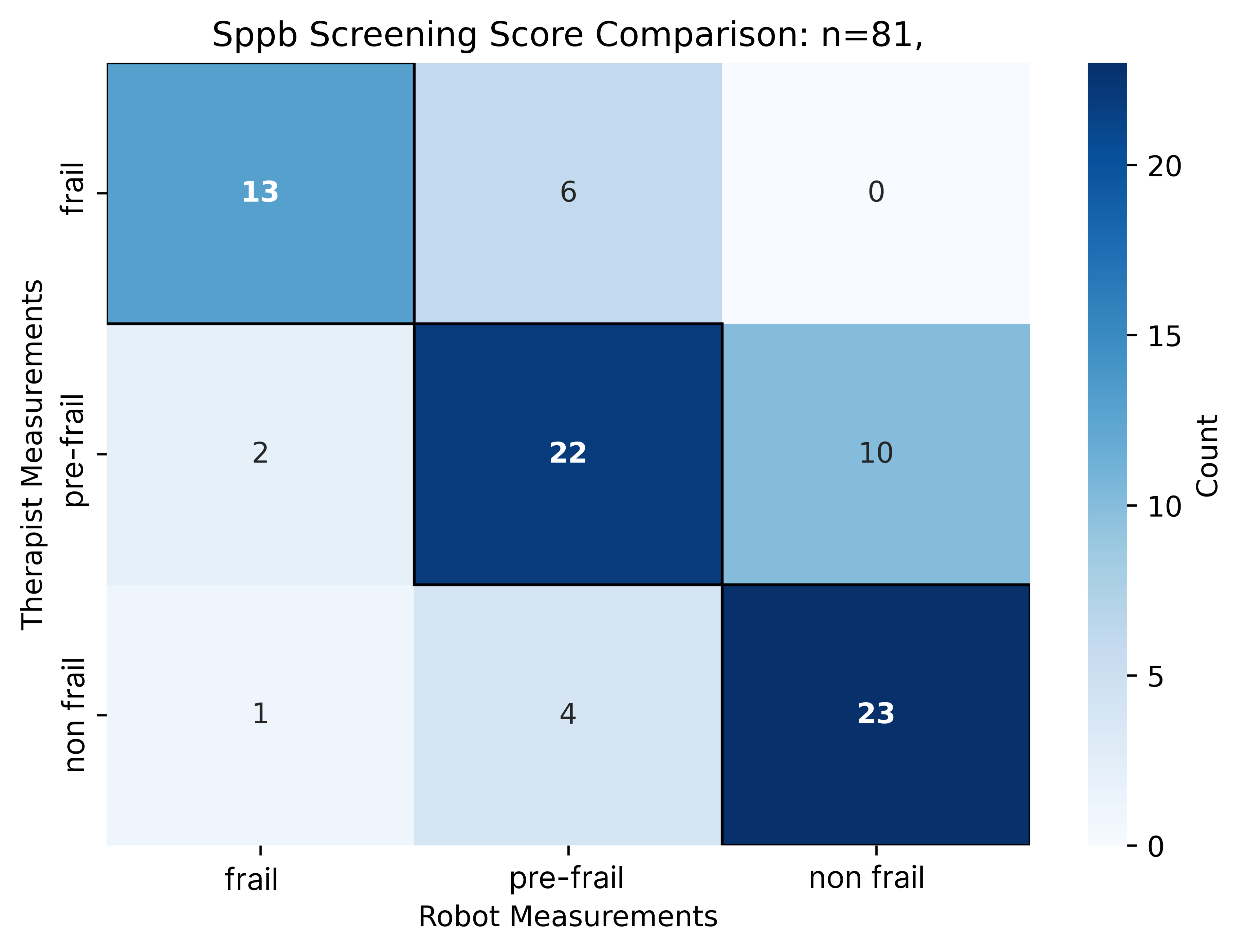}
        \caption{Screening SPPB}
        \label{fig:robot_therapist_sppb_screening_score}
    \end{subfigure}
    \caption{Confusion matrices for the test scores obtained by the robot and the therapist. The correct decisions in the confusion matrices are marked as black squares on the diagonal. Full SPPB represents the total score of the SPPB, while Screening SPPB indicates screening into frail, pre-frail, and non-frail categories.}
    \label{fig:robot_therapist_scores}
\end{figure*}

\begin{figure*}[t]
                \centering
                \begin{subfigure}{0.33\linewidth}
                    \centering
                    \includegraphics[width=\linewidth]{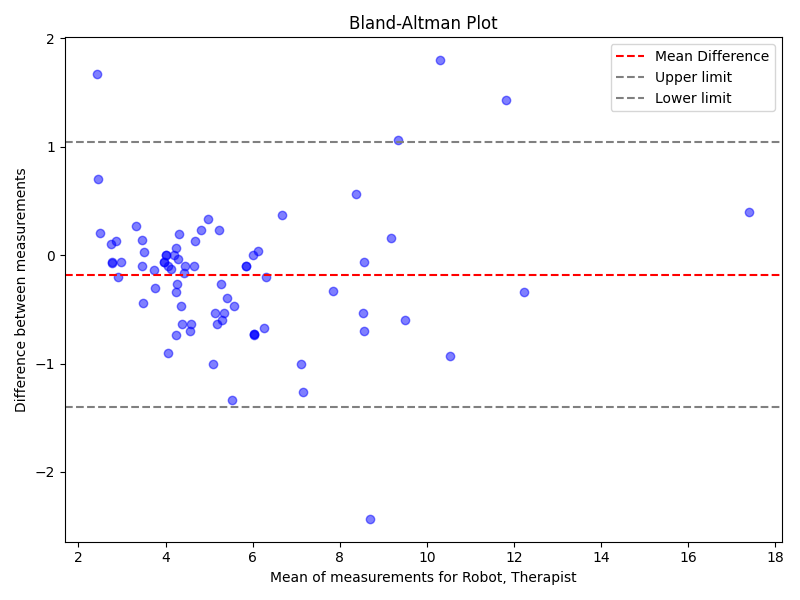}
                    \caption{Gait Speed Time (s)}
                    \label{fig:robot_therapist_gait_speed_velocity}
                \end{subfigure}\hfill
                \begin{subfigure}{0.33\linewidth}
                    \centering
                    \includegraphics[width=\linewidth]{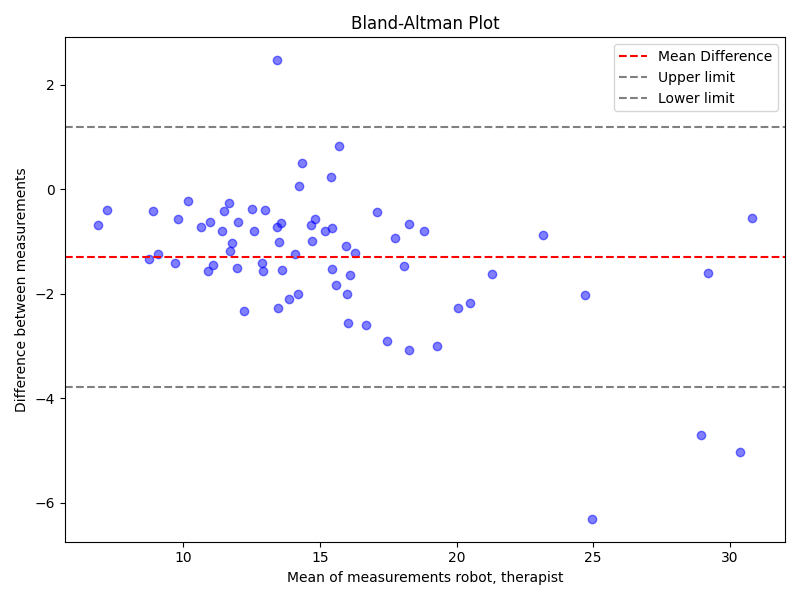}
                    \caption{Chair Stand Time (s)}
                    \label{fig:robot_therapist_sit_stand_time}
                \end{subfigure}\hfill
                \begin{subfigure}{0.33\linewidth}
                    \centering
                    \includegraphics[width=\linewidth]{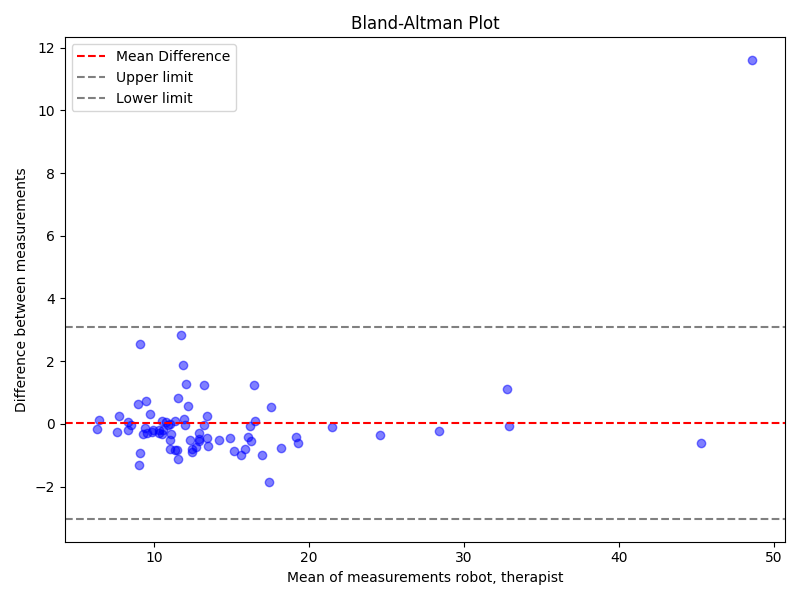}
                    \caption{TUG Time (s)}
                    \label{fig:robot_therapist_tug_time}
                \end{subfigure}\hfill
                \caption{Bland-Altman plots comparing test metrics (time and velocity) between the robot and the therapist. The dashed red line indicates the mean bias (average difference), and the dashed grey lines indicate the 95\% limits of agreement.}
                \label{fig:robot_therapist_times}
            \end{figure*}

            Table~\ref{tab:comparison_tests_robot_therapist} presents the results of the comparison between the robot and the therapist for both SPPB and TUG test times and scores. 
            The first four rows are dedicated to the scores of the full and individual SPPB tests. The last three rows report the test completion times for the Gait Speed, Chair Stand, and TUG assessments. The Standing Balance test time is excluded, as the timer stops when the participant maintains balance for 10 seconds. Consequently, most participants achieve a value of 10 seconds, whereas others differ slightly, which limits the feasibility of meaningful regression or comparison. The confusion matrices for the different test scores are shown in Fig.~\ref{fig:robot_therapist_scores}, while Fig.~\ref{fig:robot_therapist_times} presents the Bland-Altman plots, which visualise the agreement between the robot and therapist test times by plotting their differences against their averages. In both figures, the measurements show strong agreement between the robot and the therapist, with the confusion matrices falling close to the diagonal identity line and the Bland-Altman plot showing a small bias with narrow limits of agreement.

            \begin{table}[t]
                \centering
                \resizebox{1\columnwidth}{!}{%
                \begin{tabular}{|l|l|l|l|l|l|}
                    \hline
                    Metric                 & \makecell{Statistic \\ Variable} & \makecell{Number \\ Samples} & Value (IC 95\%) & Error ($\pm$SD) \\ \hline
                    SPPB score    & $\kappa$           & 81             & \textbf{0.67} (0.57, 0.76) & --- \\ \hline
                    SPPB screening    & $\kappa$           & 81             & \textbf{0.63} (0.48, 0.76) & --- \\ \hline
                    Gait Speed score       & $\kappa$           & 81             & \textbf{0.69} (0.51, 0.82) & --- \\ \hline
                    Chair Stand score        & $\kappa$           & 81             & \textbf{0.80} (0.72, 0.87) & --- \\ \hline
                    Standing Balance score & $\kappa$           & 81             & 0.39 (0.19, 0.58) & --- \\
                    \hline
                    Gait Speed time (s)        & ICC                & 76             & \textbf{0.97} (0.94, 0.99) & -0.18 ($\pm$0.62) \\ \hline
                    Chair Stand time (s) & ICC                & 69             & \textbf{0.94} (0.89, 0.97) & -1.3 ($\pm$1.27) \\ \hline
                    TUG time  (s)       & ICC                & 75             & \textbf{0.98} (0.95, 1.00) & 0.04 ($\pm$1.56) \\ \hline
                \end{tabular}
                }
            \caption{Comparison between the robot and the therapist measures of the SPPB and TUG tests.}
            \label{tab:comparison_tests_robot_therapist}
            \end{table}

            The results provide evidence that the robot estimates the SPPB and TUG test times and scores with high consistency relative to the therapist’s measurements.
            The only notable exception is the Standing Balance score, which exhibits poor reliability (see Fig.~\ref{fig:robot_therapist_standing_balance_score}). This discrepancy arises from factors not currently accounted for by the robot. As described in Sec.~\ref{sec:protocol}, a chair was placed close to the participant during the Standing Balance test for safety purposes. When a participant held onto the chair, the therapist classified the test as failed, whereas the robot could not detect it and proceeded to the next test. This occurred in 20 of the 81 participants, which significantly affected the results.            
            Conversely, the robot occasionally detected a failure when the participant had correctly performed the balance test, thereby affecting the reliability of the robot's scoring. During experimentation, it was observed that participants wearing dark or black trousers or skirts introduced additional noise in the detection of lower-limb joints (ankle, knee, and hip). This noise negatively impacted the imbalance detection algorithm, causing false imbalance detections. 
            
            More generally, it is important to note that test scores are derived from the completion time of each assessment. Therefore, discrepancies between the agreement in the Gait Speed time and the corresponding Gait Speed score are explained by the fact that several completion times were close to the threshold defining a change in score level.

            Regarding the number of participants included in each test, it can be observed that the rows corresponding to test scores contain 81 samples, whereas those corresponding to test times include fewer samples. This difference arises because, in certain cases, participants were unable to perform specific tests. In such cases, the robot assigned a score of 0. However, these trials were excluded from the time analysis since no valid test duration could be recorded.

            These results support our first hypothesis (H1a), except for the Standing Balance test, for which there are potential improvements for future work.

        \subsection{H1b: Agreement Between Robot and Reference Instruments}\label{sec:exp_times_gold_standards_robot}

\begin{figure*}[t]
                \centering
                
                \begin{subfigure}{0.24\linewidth}
                    \centering
                    \includegraphics[width=\linewidth]{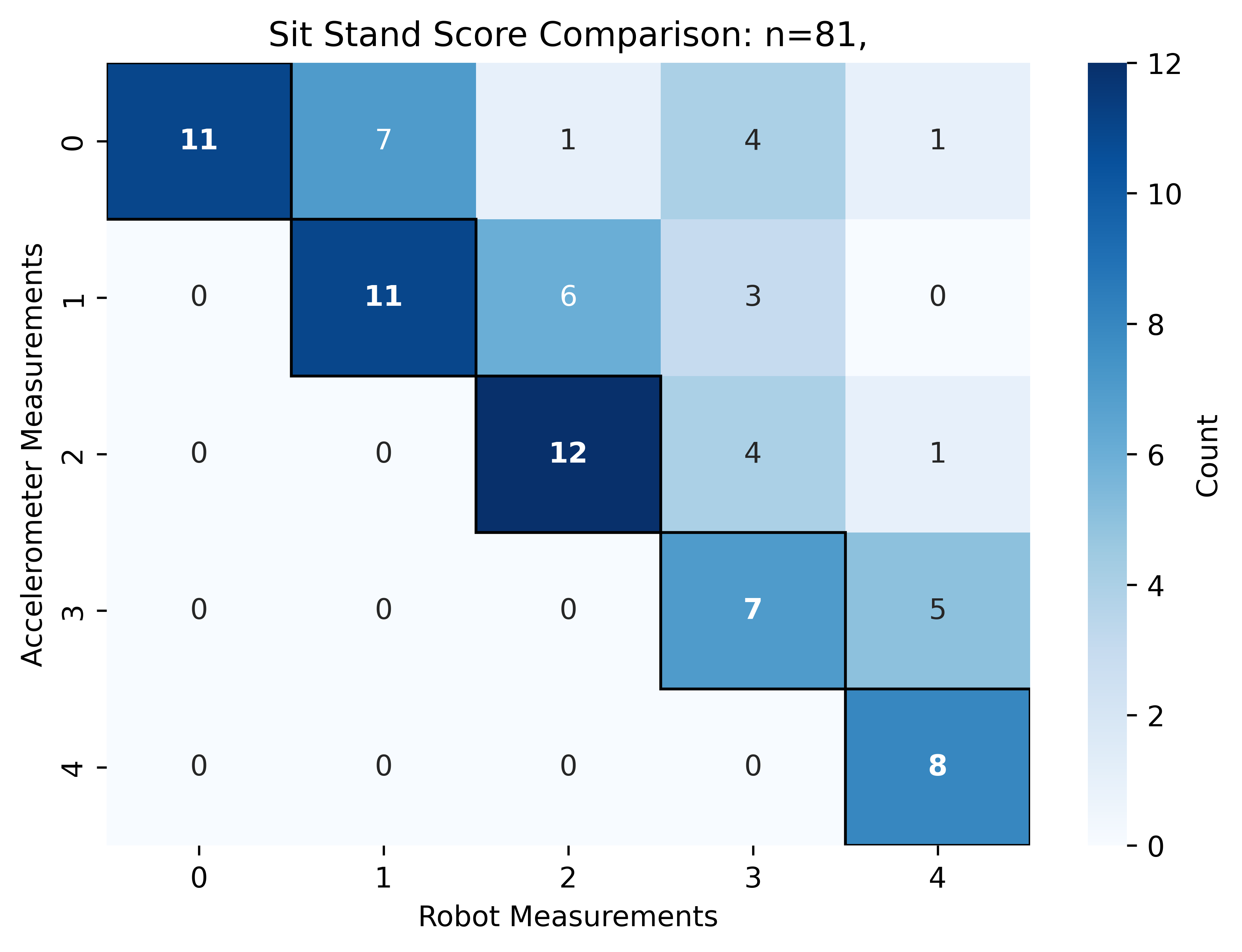}
                    \caption{Chair Stand}
                    \label{fig:robot_accelerometer_sit_stand_score}
                \end{subfigure}\hfill
                \begin{subfigure}{0.24\linewidth}
                    \centering
                    \includegraphics[width=\linewidth]{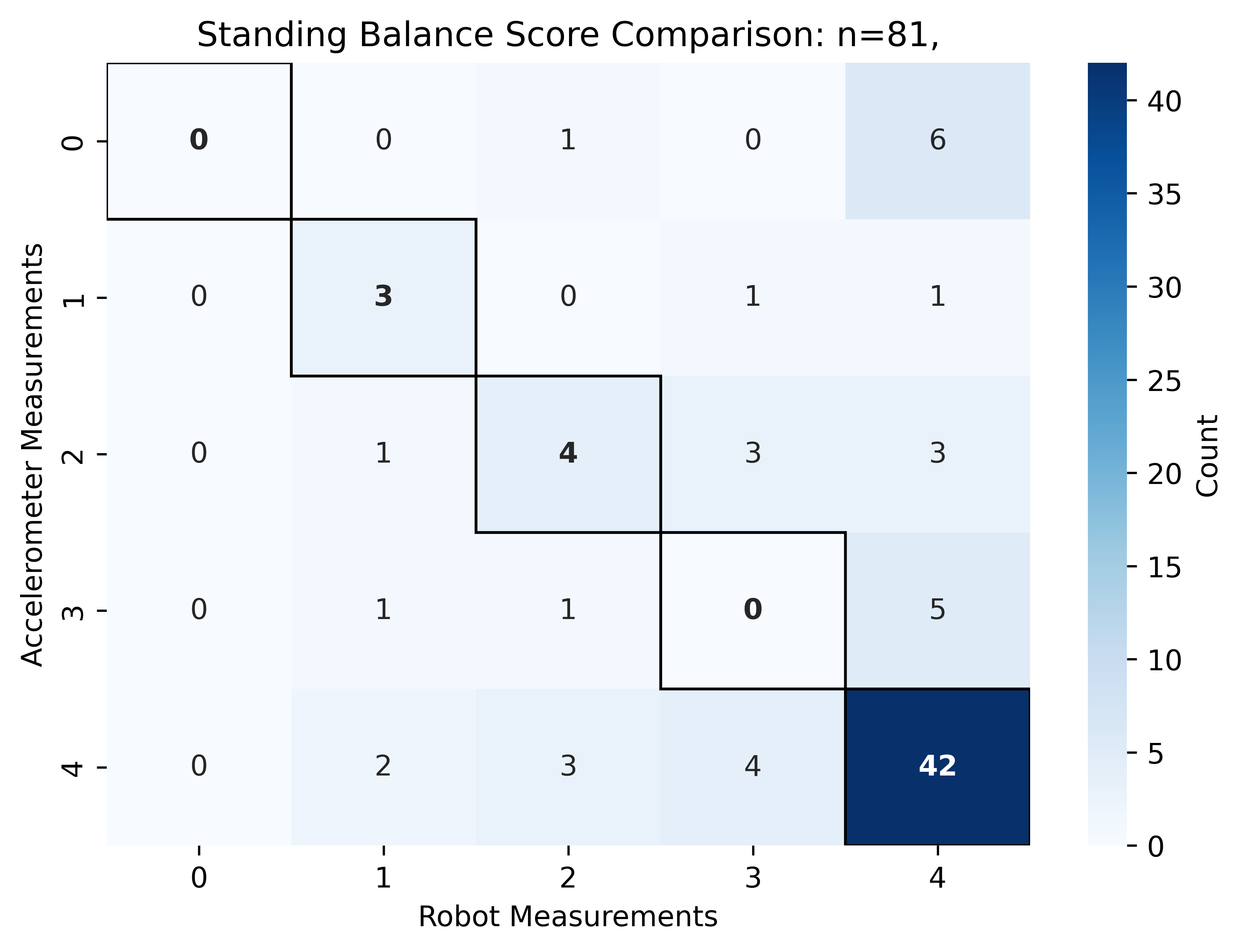}
                    \caption{Standing Balance}
                    \label{fig:robot_accelerometer_standing_balance_score}
                \end{subfigure}\hfill
                \begin{subfigure}{0.24\linewidth}
                    \centering
                    \includegraphics[width=\linewidth]{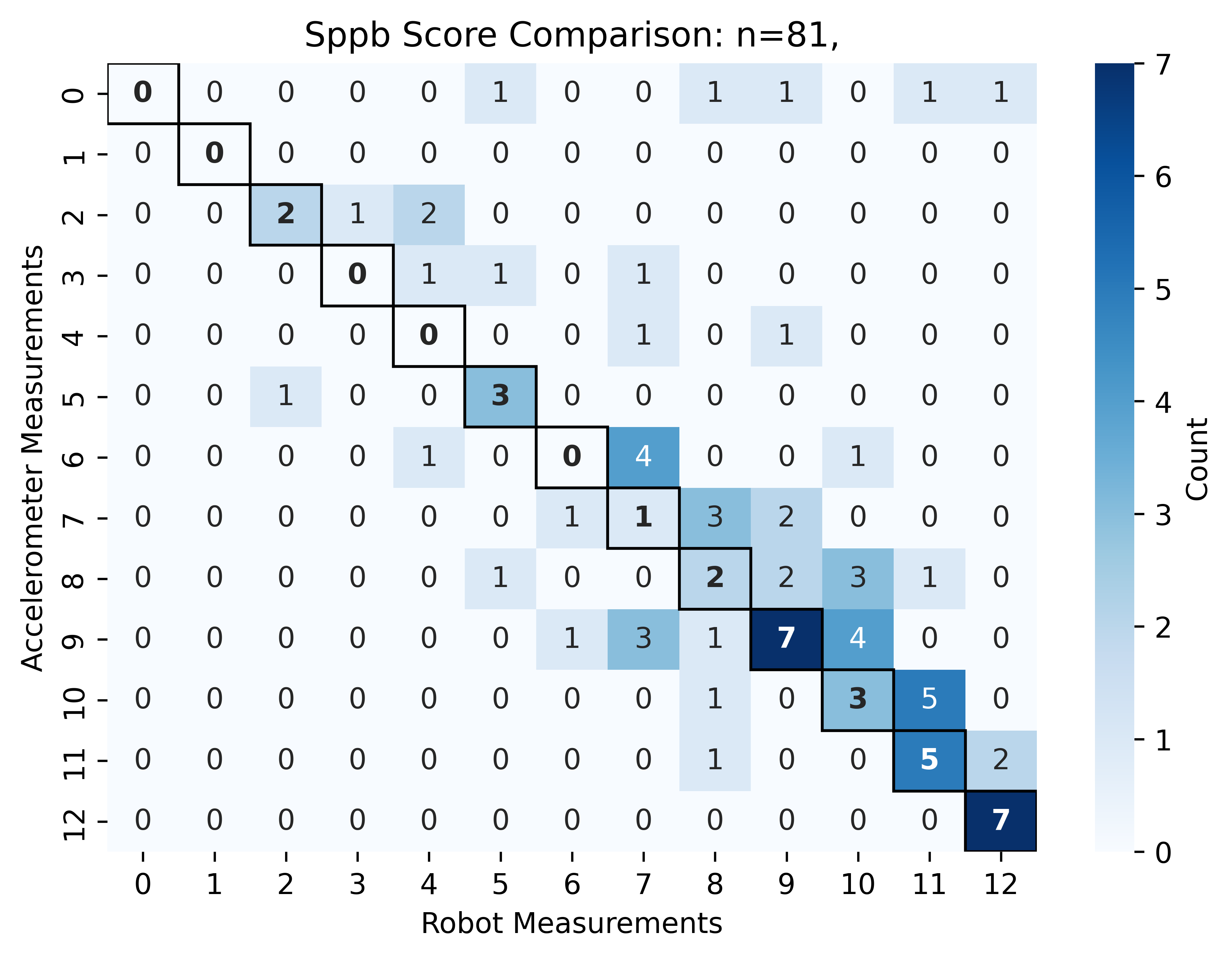}
                    \caption{Full SPPB}
                    \label{fig:robot_accelerometer_sppb_score}
                \end{subfigure}\hfill
                \begin{subfigure}{0.24\linewidth}
                    \centering
                    \includegraphics[width=\linewidth]{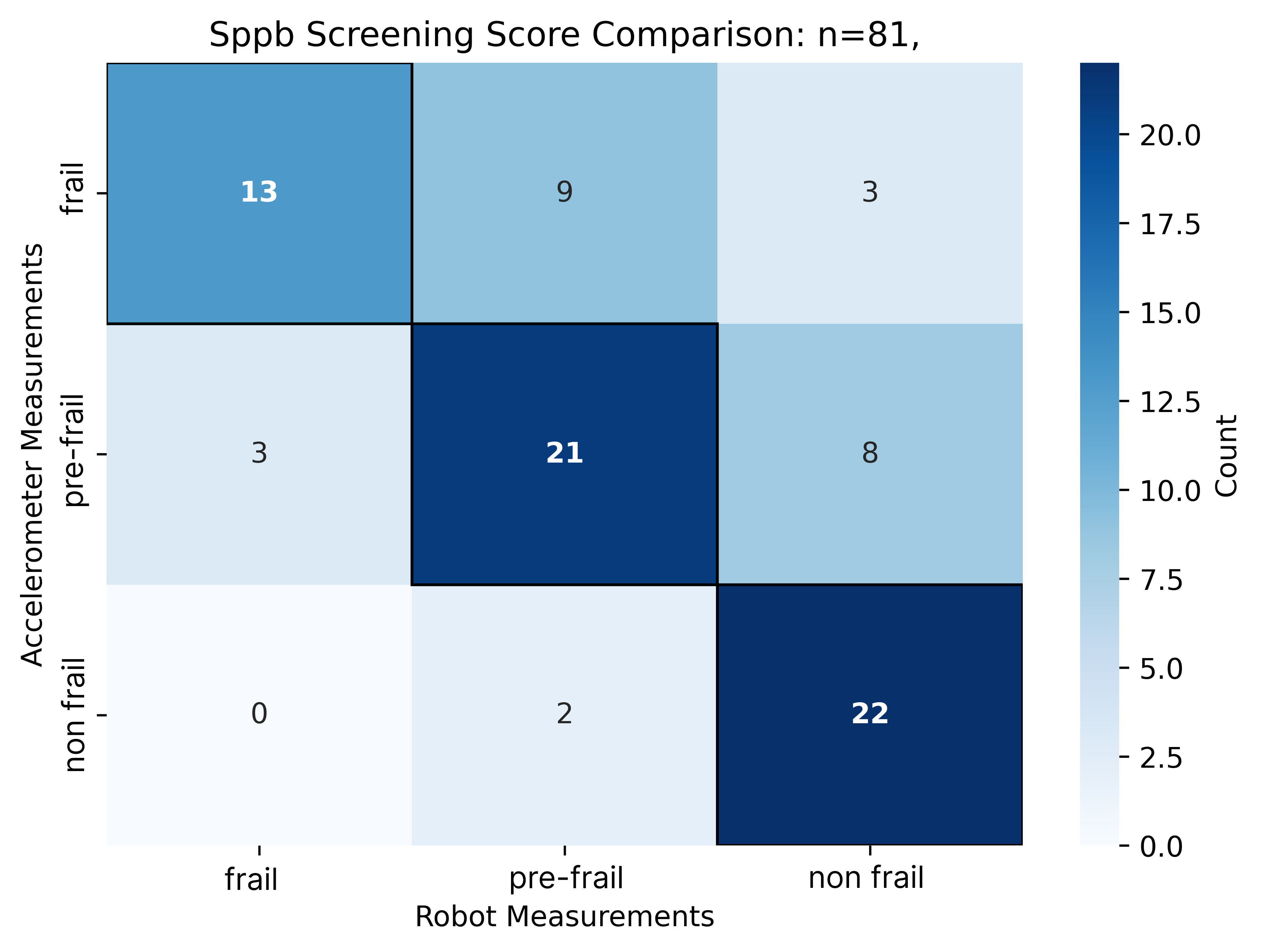}
                    \caption{Screening SPPB}
                    \label{fig:robot_accelerometer_sppb_screening_score}
                \end{subfigure}
                
                \caption{Confusion matrices for the test scores obtained by the robot (horizontal axis) and the IMU (vertical axis). The correct decisions in the confusion matrices are marked as black squares on the diagonal of the matrices.}
                \label{fig:robot_accelerometer_sppb_scores}
            \end{figure*}

\begin{figure*}[t]
                \centering
                
                \begin{subfigure}{0.33\linewidth}
                    \centering
                    \includegraphics[width=\linewidth]{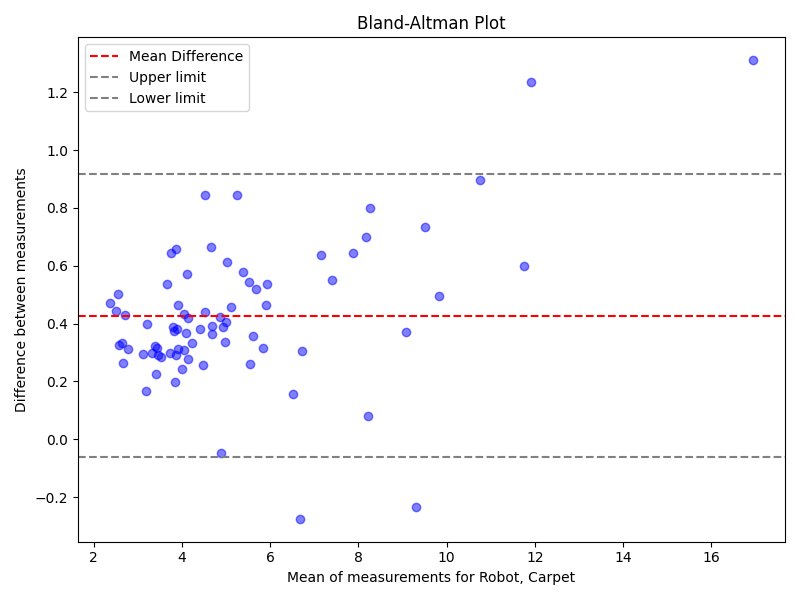}
                    \caption{Gait Speed Time - Walkway Mat}
                    \label{fig:robot_carped_gait_speed_time}
                \end{subfigure}\hfill
                \begin{subfigure}{0.33\linewidth}
                    \centering
                    \includegraphics[width=\linewidth]{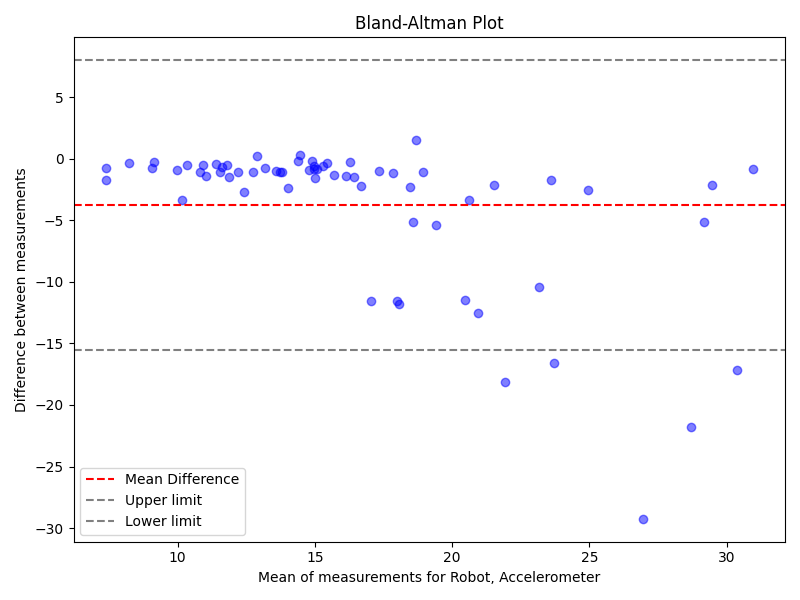}
                    \caption{Chair Stand Time}
                    \label{fig:robot_accelerometer_sit_stand_time}
                \end{subfigure}\hfill
                \caption{Bland-Altman plots comparing test metrics (velocity and time) between the robot and the reference instruments. The dashed red line indicates the mean bias (average difference), and the dashed grey lines indicate the 95\% limits of agreement.}
                \label{fig:robot_accelerometer_times}
            \end{figure*}

            Table~\ref{tab:comparison_tests_robot_gold_standards} presents the results of the comparison between the robot and the reference instruments, for both SPPB and TUG test times and scores. It is important to note that the Gait Speed scores and timings obtained from the IMU are determined by the therapist's manual start and stop during test recordings, and therefore, they are not included in this analysis. 
            For the SPPB total score, we include those measurements obtained by the therapist, therefore introducing a bias in the final score. If any discrepancy appears in the SPPB score, it will be due to the other tests (Chair Stand and Standing Balance).  
            The TUG time is also not included since the IMU's dedicated software for such tests did not compute it. A column in Table~\ref{tab:comparison_tests_robot_gold_standards} reports which reference instrument is used for computing each metric. The confusion matrices for the various test scores are presented in Fig.~\ref{fig:robot_accelerometer_sppb_scores}. Notably, the robot occasionally assigned a high SPPB score despite the IMU recording a value of zero, highlighting substantial disagreement in those instances. Fig.~\ref{fig:robot_accelerometer_times} displays Bland-Altman plots for Gait Speed time, comparing the walkway mat and the robot, and for Chair Stand Time, comparing the IMU and the robot. In the Gait Speed analysis, the comparisons fall within the ranges, indicating that the robot produces clinically acceptable estimates of gait speed.
            For the Chair Stand task, a clear trend of increasing disagreement is observed, with the robot consistently measuring lower times than the IMU.

            \begin{table}[t]
            \resizebox{1\columnwidth}{!}{%
                \begin{tabular}{|l|l|l|l|l|l|l|}
                \hline
                Metric                 & \makecell{Reference \\ Instrument} & \makecell{Statistic \\ Variable} & \makecell{Number \\ Samples} & Value (IC 95\%) & Error ($\pm$SD) \\  \hline
                SPPB score    & IMU  & $\kappa$           & 81             & 0.55 (0.42, 0.67) & --- \\ \hline
                SPPB screening    & IMU  & $\kappa$           & 81             & 0.59 (0.44, 0.73) & --- \\ \hline
                Chair Stand score        & IMU  & $\kappa$           & 81             & \textbf{0.62} (0.49, 0.73) & --- \\ \hline
                Standing Balance score & IMU  & $\kappa$           & 81             & 0.28 (0.11, 0.47)  & --- \\ \hline
                Gait Speed time ($s$)       & WM    & ICC                & 76             & \textbf{0.98} (0.97, 0.99) & 0.43 ($\pm$0.25)\\ \hline
                Chair Stand time ($s$)        & IMU  & ICC                & 65             & 0.5 (0.23, 0.76) & -3.75 ($\pm$5.99) \\ \hline
                
                \end{tabular}
                }
            \caption{Comparison between the robot and the reference instrument measures of the SPPB test. In the Reference Instruments column, WM accounts for the Walkway Mat.}
            \label{tab:comparison_tests_robot_gold_standards}
            \end{table}

            The results obtained from this comparison yield several noteworthy insights. First, the agreement between the robot’s measurements and those obtained from the IMU lies between the moderate and substantial thresholds, indicating a potential for the robot to perform frailty assessments. Second, the Chair Stand score derived from the IMU demonstrates substantial agreement with that obtained by the robot. However, a discrepancy is observed in the Chair Stand test time (see Fig.~\ref{fig:robot_accelerometer_sit_stand_time}), where the agreement is poor when using the IMU. This difference arises because, in some cases, the IMU failed to correctly detect stand-up or sit-down actions, preventing proper identification of test completion. A similar issue occurred in the Standing Balance test (see Fig.~\ref{fig:robot_accelerometer_standing_balance_score}), where the IMU occasionally detected false imbalances. 
            Finally, the comparison between the walkway mat and the robot shows excellent agreement, further confirming the robot’s suitability for accurately measuring Gait Speed performance. 

            Table~\ref{tab:comparison_accelerometer_therapist} shows the agreement in test scores between the IMU and the therapist. By doing this comparison, it is possible to infer whether any of the measurement tools (therapist, robot, and IMU) are misaligned. 
            On the one hand, we observe that the scores, when compared with the robot, differ between the IMU and the therapist. The SPPB score and the Chair Stand score, as well as the Chair Stand time, are higher for the robot-therapist than for the therapist-IMU. 
            Similarly to the previous table, the Gait Speed test is not included in the table since the times are identical (obtained by the therapist).
            Nevertheless, all measurements present a great agreement between the IMU and the therapist.

            These results support hypothesis H1b except for the Gait Speed and the Standing Balance scores.  A detailed analysis of individual tests revealed several false detections of imbalances, both in the robot and the IMU.

            \begin{table}[t]
            \resizebox{\linewidth}{!}{%
                \begin{tabular}{|l|l|l|l|l|}
                \hline
                Metric                 & \makecell{Statistic \\ Variable} & \makecell{Number\\ Samples} & Value (IC 95\%) & Error ($\pm$SD) \\ \hline
                SPPB score             & $\kappa$              & 81             & \textbf{0.76} (0.62, 0.85) & --- \\ \hline
                SPPB screening             & $\kappa$              & 81             & \textbf{0.76} (0.63, 0.86) & --- \\ \hline
                Standing Balance score & $\kappa$              & 81             & \textbf{0.73} (0.56, 0.85) & --- \\ \hline
                Chair Stand score      & $\kappa$              & 81             & \textbf{0.71} (0.59, 0.81) & --- \\ \hline
                Chair Stand time ($s$)        & ICC                & 66             & \textbf{0.79} (0.47, 0.98) & 2.48 ($\pm$5.46) \\ \hline
                \end{tabular}
                }
                \caption{Agreement between the IMU and the therapist}
                \label{tab:comparison_accelerometer_therapist}
            \end{table}

         \subsection{H2: Additional Frailty-Related Metrics}\label{sec:exp_additional_gold_standards_robot}

            Table~\ref{tab:comparison_additional_metrics} presents the results of the comparison of additional frailty-related metrics between the robot and the reference instruments. The gait-related metrics were obtained using the walkway mat, while the hip-related metrics were derived from the IMU in the Chair Stand and Standing Balance tests. As all metrics are quantitative, the ICC was used to assess the agreement between the corresponding measurements. The additional frailty-related metrics are computed only in the SPPB test and for the participants who completed the related test.

            \begin{table*}[t]
             \centering
             \resizebox{0.9\linewidth}{!}{%
                \begin{tabular}{|l|l|l|l|l|l|}
                    \hline
                    Test                      & Metric                 & \makecell[l]{Gold \\ Standard} & \makecell{Number \\ Samples} & Value (IC 95\%) & Error ($\pm$SD) \\ \hline
                    Gait Speed                & Left Step Length ($cm$)      & Walkway Mat    & 76             & \textbf{0.84} (0.74, 0.90) & 6.5 ($\pm$4.72) \\ \hline
                    Gait Speed                & Right Step Length ($cm$)     & Walkway Mat    & 76             & \textbf{0.93} (0.88, 0.97) & -1.25 ($\pm$5.01) \\ \hline
                    Gait Speed                & Left Stride Length ($cm$)    & Walkway Mat    & 76             & \textbf{0.96} (0.93, 0.97) & 5.85 ($\pm$5.31) \\ \hline
                    Gait Speed                & Right Stride Length ($cm$)   & Walkway Mat    & 76             & \textbf{0.95} (0.91, 0.97) & 6.27 ($\pm$6.3) \\ \hline
                    Gait Speed                & Left Stride Velocity ($cm/s$)  & Walkway Mat    & 76             & \textbf{0.94} (0.88, 0.97) & 7.18 ($\pm$9.51) \\ \hline
                    Gait Speed                & Right Stride Velocity ($cm/s$) & Walkway Mat    & 76             & \textbf{0.94} (0.89, 0.98) & 6.81 ($\pm$10.27) \\ \hline
                    Chair Stand               & Inclination Ascension ($^\circ$) & IMU  & 66             & -0.04 (-0.10, 0.00) & -55.30 ($\pm$22.43)\\ \hline
                    Chair Stand               & Power Ascension ($W$)       & IMU  & 66             & 0.30 (0.18, 0.51) & 39.72 ($\pm$53.11)\\ \hline
                    Chair Stand               & Acceleration Ascension ($m/s^2$) & IMU  & 66             & 0.07 (0.01, 0.20) & 1.66 ($\pm$3.31) \\ \hline
                    Chair Stand               & Acceleration Descension ($m/s^2$) & IMU  & 66             & 0.22 (-0.09, 0.52) & -0.81 ($\pm$0.53) \\ \hline
                    Standing Balance Together & Eccentricity           & IMU  & 73             & 0.15 (-0.02, 0.40) & -0.07 ($\pm$0.17)\\ \hline
                    Standing Balance Together & Stability ($mm^2$)                  & IMU  & 73             & 0.01 (-0.07, 0.28) & -1.86 ($\pm$21.92)\\ \hline
                    Standing Balance Together & Direction ($^\circ$)             & IMU  & 73             & 0.08 (-0.04, 0.21) & -42.20 ($\pm$36.07)\\ \hline
                    Standing Balance Semi     & Eccentricity           & IMU  & 69             & -0.04 (-0.30, 0.26) & 0.028 ($\pm$0.21) \\ \hline
                    Standing Balance Semi     & Stability ($mm^2$)                  & IMU  & 69             & 0.42 (0.08, 0.67) & -1.00 ($\pm$5.19)\\ \hline
                    Standing Balance Semi     & Direction ($^\circ$)             & IMU  & 69             & -0.02 (-0.19, 0.10) & -29.48 ($\pm$42.73) \\ \hline
                    Standing Balance Tandem   & Eccentricity           & IMU  & 53             & -0.13 (-0.35, 0.17) & -0.07 ($\pm$0.31) \\ \hline
                    Standing Balance Tandem   & Stability ($mm^2$)                  & IMU  & 53             & 0.41 (0.10, 0.71) & -0.033 ($\pm$8.11)\\ \hline
                    Standing Balance Tandem   & Direction ($^\circ$)             & IMU  & 53             & -0.04 (-0.21, 0.10) & -39.81 ($\pm$52.69)\\ \hline
                \end{tabular}
                }
                \caption{Comparison between the robot and the reference instruments' measures of the additional metrics of the SPPB tests.}
                \label{tab:comparison_additional_metrics}
            \end{table*}
    
            The results indicate excellent agreement between the measurements obtained by the robot and those recorded using the walkway mat. The evaluated metrics include step length, stride length, and velocity for both feet. Although the camera was positioned perpendicular to the participant’s trajectory on the right side, resulting in an occasional occlusion of the left leg on the right, the robot achieved measurements that closely matched those of the walkway mat. These findings confirm the robot’s capability to assess gait-related parameters accurately.

            In contrast, the comparison between the robot and the IMU showed generally poor agreement across most metrics, highlighting discrepancies between the two measurement systems. A likely explanation is that the additional metrics captured by the IMU are derived from a single IMU in a particular position, and both measuring tools have different noise and filters in their measurements. In the Standing Balance test, where participants must remain as still as possible, this noise can significantly affect measurement accuracy. Similarly, in the Chair Stand test, the inclination measurement depends on the precise placement of the IMU on the participant’s back, while acceleration and power are computed from a single reference point, reducing overall reliability. Also, we interpret angular metrics (e.g., direction) in an exploratory manner, while deferring to future work the study involving circular statistical metrics.

            These results validate the second hypothesis of this study only for the gait-related metrics, not for the metrics obtained from the Standing Balance and Chair Stand tests.

        \subsection{H3: Robot Autonomy During the Assessments}\label{sec:robot_autonomy}

            During the experimental sessions, the presence of a therapist in the room likely encouraged some participants to rely on the therapist to resolve doubts during assessments.

            Only one participant did not understand how to perform a test correctly: during the Gait Speed assessment, the participant walked by placing the heel of the moving foot against the tip of the other. Another participant required further clarification of the robot's physical test instructions. Five participants walked during the Standing Balance test. We hypothesise that, because the therapist explained all tests prior to the robot-led assessment, these participants retained the expectation that walking was required once standing. Two participants had difficulty hearing the robot’s instructions due to the use of hearing aids and mild echo in the empty experimental room.  

            Two participants were confused by the marks on the floor, particularly the blue cross indicating the turning point in the TUG test, and stopped walking upon reaching it. One participant was unable to see the floor markings clearly. Two participants lost count of the number of chair stand repetitions and asked the therapist how many remained. One participant did not press the Start button before beginning the TUG test. In one instance, the robot failed to announce ``You can start” during the Chair Stand test, prompting therapist intervention to signal the start of the task.

            Furthermore, one participant recently had an ictus, which induced some cognitive impairment and difficulty understanding the robot instructions. Therefore, a dedicated therapist had to intervene and provide supervision and guidance. Similarly, a participant with dementia required the same intervention. 

            In total, therapist intervention was required for 17 out of 81 participants. Nevertheless, most of these issues could be mitigated through more flexible and adaptive robot behaviour. For example, the robot could physically move toward floor markings to demonstrate where participants should walk, offer optional, more detailed explanations, or adapt its interaction style based on the user’s individual needs and conditions. Furthermore, the therapist's authority was implicitly established when welcoming participants and explaining the protocol, which facilitated their reliance on them for asking simple doubts.

\section{Limitations and Outlook}

        The presented system has proven to be a powerful tool for robots to assess older adults' frailty in healthcare environments. However, it is important to acknowledge certain limitations that should be addressed in future work.
        
        Firstly, during the Standing Balance test, those participants who were already familiar with the procedure positioned their feet according to the visual instructions before the countdown began, thereby prolonging the test beyond the required time. For frail individuals, this additional effort may induce fatigue, potentially influencing the test outcomes and increasing the risk of imbalance or falls. Secondly, participants with limited balance often required the presence of a support for safety. Future implementations should ensure that the testing environment accounts for these safety needs and that the robot is capable of detecting when participants hold onto such supports, as doing so should terminate the tests. Thirdly, black trousers and skirts introduced noise in the detection of knee and ankle keypoints, which are critical for identifying sitting and standing actions, balance events, and gait information.  
        In addition, hip keypoint detection was less accurate for participants with wider hip proportions, affecting the algorithm's ability to determine stand-ups and sit-downs reliably. Fourthly, due to a limitation on personal availability, the therapist had to operate the reference instruments and the timer simultaneously, which could induce a small dependency between time measures. Future work could use a synchronised trigger.

        Nevertheless, the framework offers advantages with respect to the reference instruments' procedures. The IMU frequently lost connection with its paired device, interrupting data collection. Before each interaction, we had to reconnect the IMU to avoid losing connection between tests, which happened on 6 occasions, provoking data loss. Additionally, in other studies involving comparisons between an IMU and robot clinical measures, they found that the IMU sometimes provides wrong measures (i.e., in ~\cite{Calabrese_IJSR25}). 
        Unlike the reference instruments, the robot does not require user-specific calibration. For example, the walkway mat must be calibrated for each participant based on height and weight. Furthermore, the robot eliminates the need for intrusive or stationary apparatus: the walkway mat must be installed in a dedicated space and operated by a therapist, and the IMU must be attached to the participant’s back using a belt. The robot, by contrast, guides users through the assessment and could be further developed to assist throughout the clinical workflow: locating patients, accompanying them to the evaluation area, and guiding them to the doctor’s office afterwards. 

        In general, greater safety measures are required compared to conventional assessments, as the patient would interact with the robot independently, removing the safety from direct human supervision. Although this study was conducted in a controlled environment where such risks were minimised, future developments should include enhanced safety protocols. Examples include allowing the robot to skip tests if a participant feels unsafe, or performing brief preliminary trials (e.g., a single chair stand) to evaluate readiness and minimise risk before proceeding with the full assessment.

\section{Conclusions}

    In this work, we presented a framework that enables robots to autonomously perform frailty and fall-risk assessments for older adults under the supervision of a human therapist. The framework, based on Behaviour Trees, provides efficient and structured guidance throughout the evaluation, using simple verbal cues to assist participants. Interaction between the robot and the participants was facilitated through a tactile interface displayed on an integrated screen.

    The experimental results support our initial research hypotheses, showing a strong agreement between the test measurements obtained by the therapist and those generated by the robot, as well as good agreement between the robot and the reference instruments for both SPPB scores and completion times. Moreover, the robot’s gait-related measurements showed excellent correspondence with those derived from the walkway mat.

    These findings indicate that robots equipped with the proposed framework have strong potential as reliable tools for clinical frailty and risk of falling assessment. With the incorporation of appropriate safety protocols, such robotic systems could be deployed in hospitals or clinical settings to perform preliminary screenings prior to medical consultations, thereby reducing clinicians’ workload and providing objective, consistent measurements to support diagnostic decision-making.

\section*{References}

\end{document}